\documentclass[mnsc,nonblindrev]{informs3}

\OneAndAHalfSpacedXI % current default line spacing
\usepackage{natbib}
 \bibpunct[, ]{(}{)}{,}{a}{}{,}%
 \def\bibfont{\small}%

\EquationsNumberedThrough 
\MANUSCRIPTNO{MS-MKG-20-03054}

\usepackage{graphics}
\usepackage{xcolor}
\usepackage{bbm}

\usepackage{caption}
\usepackage{subcaption}
\usepackage{hyperref}
\hypersetup{
     colorlinks   = true,
     allcolors    = [rgb]{0.639,0.122,0.234}
}

\usepackage{epigraph}

\usepackage{indentfirst}
\usepackage{authblk}
\usepackage{verbatim}
\usepackage[T1]{fontenc}
\usepackage[utf8x]{inputenc}
\usepackage{hologo}
\usepackage{textcomp}
\usepackage{hanging}

\usepackage{amsmath}
\usepackage{amssymb}
\usepackage{amsfonts}
\usepackage{amsfonts}
\usepackage{booktabs}
\usepackage{siunitx}
\usepackage{tikz}
\usetikzlibrary{arrows.meta}
\usepackage{appendix}

\usepackage{float}
\usepackage{authblk}

\renewcommand{\thesection}{\arabic{section}}
\definecolor{cardinal}{rgb}{0.639,0.122,0.234}

\newtheorem{prop}{Proposition}
\newtheorem{define}{Definition}
\newtheorem{assump}{Assumption}

\newcommand{\E}{\mathbb{E}}

\newcommand\indep{\perp \!\!\! \perp} 
\newcommand\sgn{\text{sign}} 
\newcommand\CATE{\tau}
\newcommand\tCATE{\tilde{\tau}}

\begin{document}
\TITLE{Targeting for long-term outcomes}

\ARTICLEAUTHORS{%
\AUTHOR{Jeremy Yang}
\AFF{Harvard Business School, \EMAIL{jeryang@hbs.edu}} %, \URL{}}
\AUTHOR{Dean Eckles}
\AFF{Massachusetts Institute of Technology, \EMAIL{eckles@mit.edu}}
\AUTHOR{Paramveer Dhillon}
\AFF{University of Michigan, \EMAIL{dhillonp@umich.edu}}
\AUTHOR{Sinan Aral}
\AFF{Massachusetts Institute of Technology, \EMAIL{sinan@mit.edu}}
} % end of the block

\ABSTRACT{%
Decision makers often want to target interventions so as to maximize an outcome that is observed only in the long-term. This typically requires delaying decisions until the outcome is observed or relying on simple short-term proxies for the long-term outcome. Here we build on the statistical surrogacy and policy learning literatures to impute the missing long-term outcomes and then approximate the optimal targeting policy on the imputed outcomes via a doubly-robust approach. We first show that conditions for the validity of average treatment effect estimation with imputed outcomes are also sufficient for valid policy evaluation and optimization; furthermore, these conditions can be somewhat relaxed for policy optimization. We apply our approach in two large-scale proactive churn management experiments at \textit{The Boston Globe} by targeting optimal discounts to its digital subscribers with the aim of maximizing long-term revenue.  Using the first experiment, we evaluate this approach empirically by comparing the policy learned using imputed outcomes with a policy learned on the ground-truth, long-term outcomes. The performance of these two policies is statistically indistinguishable, and we rule out large losses from relying on surrogates. Our approach also outperforms a policy learned on short-term proxies for the long-term outcome. In a second field experiment, we implement the optimal targeting policy with additional randomized exploration, which allows us to update the optimal policy for future subscribers. Over three years, our approach had a net-positive revenue impact in the range of \$4-5 million compared to the status quo.
}%

%\KEYWORDS{deterministic inventory theory; infinite linear programming duality;
%  existence of optimal policies; semi-Markov decision process; cyclic schedule}

\KEYWORDS{long-term effect, statistical surrogate, policy learning, targeting, proactive churn management} 
\HISTORY{First version: October 2020. This version: February 2022.}

% \HISTORY{This paper was
% first submitted on April 12, 1922 and has been with the authors for
% 83 years for 65 revisions.}

\maketitle

% %\doublespacing
% \onehalfspacing

\section{Introduction}
Advertising revenues have been stagnating for newspapers in recent years.\footnote{The print advertising revenue is declining with a compound annual growth rate (CAGR) of -12.6\% from 2016-2021, while digital ads revenue is still growing at a CAGR of 2.2\%, it's not enough to compensate for the loss in print. Source: US Online and Traditional Media Advertising Outlook} As a consequence, newspapers are looking for ways to strengthen their subscription-based business model. Take \emph{The New York Times} as an example: in 2019, their total subscription revenue was twice their total advertising revenue (Figure \ref{nyt1}, \ref{nyt2}). Their CEO recently said: ``$\ldots$ we still regard advertising as an important revenue stream, but we believe that our focus on establishing close and enduring relationships with paying, deeply engaged subscribers, and the long-range revenues which flow from those relationships, is the best way of building a successful and sustainable news business''.\footnote{Source: https://www.nytimes.com/2018/02/08/business/new-york-times-company-earnings.html} Hence, to succeed in a subscription-based business model, news publishers must retain their existing subscribers and maximize their long-term values. A common approach to achieving this goal is to target existing subscribers with marketing interventions, such as price discounts or other personalized offers. 

We use news publishers as a motivating example, and it matches our empirical application. But how to optimize long-term customer outcomes by targeting interventions is a problem faced by most firms. Even more generally, decision makers in education, government, and medicine typically care about intervening for long-term outcomes such as employment, income, and survival.

``Long-term'' and ``short-term'' outcomes are fruitfully understood as defined relative to the targeting cycle. For example, if a firm runs a campaign every year, then all outcomes that are observed within a year, such as their one-year revenue, might be considered ``short-term'' because these outcomes are observed before the firm takes action (decides whom to target with what) in their next campaign. Hence, future policies can be optimized on these observed outcomes. In contrast, ``long-term'' outcomes materialize over time horizons longer than the window of opportunity for action, for example, three-year or five-year revenue, rendering the firm incapable of optimizing their next campaign based on them. So, a natural question arises: How can firms learn and implement an optimal targeting policy when the primary outcome of interest is ``long-term''?

A straightforward solution to this problem is to wait until the long-term outcome materializes and choose a policy based on the realized long-term outcome. But this implies that the firm can not learn anything in the meantime, and therefore is unable to implement updated targeting policies until years later. Another solution is to find a short-term proxy (e.g., short-term revenue) for the long-term outcome and optimize for it instead. However, this could be problematic as the proxy and the long-term outcome might not be well aligned. Hence, a policy that performs well on the proxy might not perform well in the long-run. 

In this paper we propose to use surrogates~\citep{prentice1989surrogate,vanderweele2013surrogate} to impute the missing long-term outcomes and use the imputed long-term outcomes to optimize a targeting policy. We estimate the missing long-term outcome as the expectation of the long-term outcome conditional on surrogates of that outcome in a historical dataset in which the long-term outcome was observed. Surrogate index estimators combine multiple surrogates by estimating the conditional expectation of the long-term outcome given the surrogates and using this to impute long-term outcomes~\citep{xu2001evaluation,athey2019surrogate}. Once we have the imputed long-term outcomes, we optimize the targeting policy efficiently by using a doubly-robust approach~\citep{dudik2014doubly, athey2017efficient, zhou2018offline} on the imputed long-term outcomes. We prove this surrogate-index-based approach recovers the optimal policy learned on true long-term outcomes under certain assumptions. We implement the optimal policy via bootstrapped Thompson sampling \citep{eckles2014thompson,osband2016deep} to maintain exploration so we can update and re-optimize the policy for future subscribers to allow for potential non-stationarity.

We evaluate the efficacy of our approach empirically by running two large-scale field experiments that target discounts to the digital subscribers of \emph{The Boston Globe}, a regional leader in news media. Boston Globe Media, which operates \emph{The Boston Globe} newspaper and associated websites, is facing a similar problem to many other publishers. Our goal is to learn an optimal targeting policy that treats some subscribers with certain discounts to maximize their retention and long-term revenue. Here a policy is a mapping from subscriber characteristics to offering a specific discount (or no discount, or a distribution over discounts when the policy is stochastic). In this subscriber retention context, this is also known as proactive churn management.\footnote{Here proactive simply means that the intervention (discount) happens before a churn intention is observed; by contrast, reactive churn management means that the company first waits for customers to request to cancel their subscription then offers some discount or other benefits in reaction to this in the hope of retaining them. One analogy is that the proactive approach is like diagnosing and preventing illness before a patient shows clear symptoms, and the reactive approach is like treating patients who are already ill.} To construct the surrogate index, we use the observed revenue and content consumption in the 6 months after treatment as our surrogates. We compare how well the policies learned using surrogate index perform against policies optimized directly on short-term proxies (a benchmark) or realized long-term outcomes (the ground truth). We also consider alternative selections of surrogates for the construction of surrogate index --- perhaps most importantly whether we can use less than 6 months of revenue and consumption data. We estimate that this approach increases the firm's total projected digital subscription revenue by \$4--5 million over a three-year period relative to the status quo in the two experiments.

The rest of the paper is organized as follows. In Section~\ref{related work} we review related work. The empirical context is described in Section~\ref{context}. We introduce our method in Section~\ref{method}: We first explain the imputation of the long-term outcome using the surrogate index and prove sufficient conditions for it to be valid for policy evaluation and optimization, then we describe the policy learning framework and how it is implemented. Experimental results and empirical validation of our approach are reported in Section~\ref{result}. We conclude in Section~\ref{conclusion}.

\section{Related Work} \label{related work}
Our paper builds on a large body of literature in biostatistics and medicine on surrogate outcomes (i.e., endpoints, biomarkers); see, e.g., \cite{joffe2009related} and \cite{weir2006statistical} for reviews. In clinical trials the goal is often to study the efficacy of an intervention on outcomes such as the long-term health or survival rate of patients. However, the primary outcome of interest might be very rare, only observed after years of delay, or have high variance compared with the treatment effects (e.g., a 5 or 10-year survival rate). It is common to use the effect of an intervention on surrogate outcomes as a proxy for its effect on long-term outcomes. In a seminal paper, \cite{prentice1989surrogate} argued that to be a valid surrogate, treatment and outcome have to be independent conditional on the surrogate. One intuitive way for this condition to be satisfied is if the surrogate fully mediates the treatment effect.
In practice it is hard to find a single variable that plausibly satisfies the condition \citep{freedman1992statistical}, but \cite{xu2001evaluation} showed that combining multiple surrogates to predict the outcome can be preferable to using a single surrogate because the treatment effect may operate through multiple pathways and, even when there is a single pathway, using multiple surrogates can reduce measurement error. This idea is further developed in a recent paper in econometrics \citep{athey2019surrogate}, where the combination is referred to as a \emph{surrogate index}. This literature focuses on using surrogates to identify treatment effects on long-term outcomes and, in this paper, we extend this to targeting policy optimization.

Another popular approach to modeling long-term outcomes is to posit a particular parametric generative model for the long-term outcomes. In the context of marketing, this is typically a model of customer lifetime value (CLV or LTV). CLV models are widely used in marketing for customer segmentation and targeting; see, for example, \cite{gupta2006modeling}, \cite{fader2014stochastic}, \cite{fader2015simple}, and \cite{ascarza2017marketing} for surveys. CLV is defined as the sum of discounted future revenues or profits from a customer. To calculate CLV we typically need to posit a parametric, e.g., survival function and extrapolate the survival or retention probability into the future. A recent example in the context of churn management is \citet{godinho2018target}, where a parametric survival function is used. 
One advantage of this approach is that we can apply it even when the long-term outcomes are never observed
%\footnote{The model is often assumed to be infinite horizon.}
because the prediction is based on functional form assumptions, unlike the surrogate index approach which needs access to long-term outcomes in a historical dataset; on the other hand, standard parametric CLV approachs may suffer from model misspecification. Also, the primary goal of CLV models is typically to predict outcomes, whereas the surrogate index approach focuses on learning treatment effects or optimizing policies, imputing outcomes is just a means to an end; more importantly, outcomes imputed via surrogate index have provable properties regarding treatment effect estimation \citep{athey2019surrogate} or policy learning, as developed here. %Another difference is that CLV models are usually cohort-based, whereas the surrogate index imputes outcomes at the individual level. 
Furthermore, building a CLV model may require substantial work to formalize business logic in anything but the simplest subscription businesses.
A synthesis of these approaches is also possible in that a CLV prediction, if already available, can also be used as one of the surrogates in the construction of a surrogate index.

This paper is also related to the literature on targeting policy evaluation and optimization, which has recently further developed within marketing research. \cite{hitsch2018heterogeneous} proposed an estimation method for conditional average treatment effects (CATEs) based on k-nearest neighbors (kNN) and used it for policy optimization. \cite{simester2019efficiently} showed that we can compare targeting policies more efficiently if we only compare the outcome of units on whom the policies prescribe different actions. \cite{simester2019targeting} documented non-stationarity such as covariate and concept shifts between two experiments and evaluated how robust different machine learning models used to optimize policies are to these changes in the environment. \cite{yoganarasimhan2020design} used different machine learning models to estimate CATEs and evaluated how targeting policies constructed using these models perform against each other.  In another recent work, \citet{lemmens2020managing} examine using a CLV model combined with field experimentation to optimize targeting in the policy learning framework.

Our work complements this literature by developing an approach that is novel in a few ways. First, we focus directly on targeting for long-term outcomes; outcomes used in these other works are short-term (in the sense that they are observable when we optimize and implement the policy) or extrapolation is done using a parametric CLV model.\footnote{\cite{yoganarasimhan2020design} showed in their particular case the policy learned on short-term outcome also does well on long-term outcomes, but the policy is not directly optimized on long-term outcome.} Second, we systematically add randomized exploration around the learned policy, which allows us to evaluate and update the policy for future units in case the environment changes. \cite{hitsch2018heterogeneous} and \cite{yoganarasimhan2020design} studied the problem in a static setting. \cite{simester2019targeting} did look at changes in the environment but they focused on evaluating the robustness of different machine learning models. Third, we use a doubly-robust (DR) approach \citep{dudik2014doubly} for both policy evaluation and learning in contrast to \cite{hitsch2018heterogeneous} and \cite{yoganarasimhan2020design} who used an inverse probability weighting (IPW) estimator for policy evaluation.  \citet{lemmens2020managing} introduce a specialized incremental profit based loss function that performs well in their empirical evaluation, but lacks the asymptotic efficiency results available for doubly-robust policy learning; it is also unclear how to combine this with known probabilities of treatment (i.e., design-based propensity scores) that arise in sophisticated experiments. In particular, even when probabilities of treatment are known exactly (as in our setting), DR estimators have advantages in statistical efficiency compared with IPW estimators \citep{athey2017efficient, zhou2018offline}.

Substantively, our study adds to the literature on subscriber management and proactive churn management in particular. Earlier work focused on developing better prediction algorithms to more accurately identify potential churners; \cite{neslin2006defection} provides a detailed comparison of different churn prediction models. Recently, the literature has examined causal effects of targeting interventions on churn using field experiments. For example, \cite{ascarza2018retention} and \cite{lemmens2020managing} note that firms should not target customers based on their outcome level (churn risk) but should target based on treatment effects. \cite{ascarza2016perils} showed evidence from a field experiment with a telecommunication company that proactive churn interventions can backfire and increase the churn rate in practice. They argued that this is because proactive intervention lowers customers' inertia to switch plans and increases the salience of past-usage patterns among potential churners.
Our paper contributes to this literature by proposing an experimental framework that can be applied to directly optimize targeting policies for long-term customer retention and revenues.

\section{Empirical Context} \label{context}

Founded in 1872, \emph{The Boston Globe} is the oldest and largest daily newspaper in the greater Boston area. It has won a total of 26 Pulitzer Prizes and is widely regarded as one of the most prestigious papers in the US. We ran two targeting experiments on all digital only\footnote{\emph{The Globe} also has a combined print and digital subscription. All subscribers are paying customers.} subscribers of \emph{The Boston Globe} in two experiments. While we return to the details of our experiments and analyses in Section \ref{result}, we introduce the empirical context here so as to help fix ideas as we describe the methods.

Our analysis is of a random sample of about 45,000 digital subscribers in the first experiment and 95,000 in the second. For each subscriber we observed the short-term outcome (e.g., monthly churn and revenue) and three sets of features: demographics (e.g., zip code), account activities (e.g., billing address change, credit card expiration date, complaints), and content consumption (e.g., when and what articles they read). There was only one intervention in the first experiment, which lowered the price for treated subscribers from \$6.93 per week to \$4.99 per week for 8 weeks. %Approximately 1,000 subscribers were treated in the first cohort. 
An email (Figure \ref{fig1}) was sent to all treated subscribers in August 2018 telling them that a discount had been automatically applied to their accounts. We implemented 6 interventions in the second experiment:  a thank you  email, a  \$20  gift  card, a discount to \$5.99 for 8  weeks, a discount to \$5.99 for 4  weeks, a discount to \$4.99 for 8 weeks (the same as the intervention in the first experiment), and a discount to \$3.99 for 8 weeks. 
A similar email (Figure \ref{fig2}) was sent to all treated subscribers in July 2019 with the corresponding message, and a treated subscriber had to click on a button at the bottom of the email to redeem the benefit. There was no overlap of treated subscribers between the two experiments.

\section{Methods}
\label{method}
In our application, the primary outcome of interest is long-term subscriber retention or revenue\footnote{Being a digital service, marginal costs are negligible compared with subscription revenue.}, but we do not observe these outcomes in the short-term, i.e., after the intervention in the first experiment and before we implemented the learned policy for the second experiment of customers. Hence, we use a surrogate index to address this problem. 

Our framework has two components: First, we fit a model for long-term outcomes and use the resulting surrogate index to impute long-term outcomes; second, we learn an optimal policy using the imputed long-term outcomes. In Section \ref{surrogate index} we explain the imputation and prove sufficient conditions for it to be valid for policy evaluation and optimization. In Section \ref{policy} we  describe  the  policy evaluation and optimization framework and how it is implemented. 

We first introduce the notation that we use throughout the section: Let $\pi \in \Pi$ be a targeting policy that maps from the space of unit characteristics $\mathbb{X}$ to a space of distributions (simplex) over a set of discrete actions $\mathbb{A}$; we index actions by $\{0,1,2,..,K-1\}$, where 0 is control and others are different interventions. When the policy is non-deterministic, it defines a non-degenerate probability distribution over possible actions conditional on covariates $\pi(a|x) := \mathbb{P}(A=a|X=x), \forall a \in \mathbb{A}, x \in \mathbb{X}$. When it is deterministic, it maps to a fixed action with probability 1. Depending on the action chosen, we observe the corresponding potential outcome, i.e., $Y_i = Y_i(A_i)$. 
These potential outcomes may be correlated with unit characteristics $X_i$.

The goal is to learn a policy that maximizes some average outcome $Y$ (if the goal is to minimize some average outcome $Y$, we can add a negative sign and turn it into a maximization problem).
%The average long-term outcome under a given policy $V(\pi)$ is the value of that policy: 

\begin{define}{A Policy and its Value}
\begin{gather}
\pi: \ \mathbb{X} \to \Delta(\mathbb{A})\\
V(\pi) :=  \mathbb{E} [Y_i(A_i)]
\end{gather}
\end{define}

\begin{define}{Optimal Policy}
\begin{equation}
\pi^* :=\underset{\pi \in \Pi}{{\mathrm{argmax}}} \ V(\pi)
\end{equation}
\end{define}

\subsection{Imputing a Long-term Outcome with a Surrogate Index} \label{surrogate index}

% \subsection{Notation}
% \label{setup}

We use intermediate outcomes that are observed over the short-term period following the intervention as surrogates. Intuitively, the idea is to select surrogates that capture some of the ways that the actions affect the long-term outcome; in our application, these are subscriber's content consumption and short-term revenue.
These surrogate variables are then combined with the long-term outcomes in the historical dataset to impute missing long-term outcomes for units in the experiment.

Assume we have two datasets, one from the experiment labeled $E$ and one based on historical (observational) data labeled $H$. We observe draws of the tuple $(X,A,S)$ in the experiment where $X \in \mathbb{X}$ represents units' baseline characteristics, $A \in \mathbb{A}$ is the action (i.e., treatments, interventions), and $S \in \mathbb{S}$ is the potentially vector valued set of intermediate outcomes or surrogates. Note that the long-term outcome $Y$ is unobserved in the experiment. In the historical dataset, we observe draws of the tuple $(X, S, Y)$; note that is no known, randomized intervention in this dataset (i.e., it is observational), but the long-term outcome $Y$ is observed.
We can define a surrogate index $\tilde{Y}$ for the long-term outcome $Y$ as the expectation of the long-term outcome conditional on unit covariates and surrogates in the historical dataset $H$:\footnote{One advantage of this approach is that the estimation of the conditional expectation can be treated as a supervised learning problem and can be performed using flexible non-parametric machine learning methods like XGBoost~\citep{chen2015xgboost}.}

\begin{define}{Surrogate Index}
\label{def:si}
\begin{equation}
\tilde{Y}_i:= \E_H[Y_i|S_i,X_i]    
\end{equation}
\end{define}

\noindent Under Assumptions 1--3 stated below, a central result in \cite{athey2019surrogate} is that the average treatment effect (ATE) on $\tilde{Y}$ recovers the ATE on long-term outcome $Y$. That is, by constructing the surrogate index we can identify and feasibly estimate the ATE on some long-term outcomes without having to wait until they are observed. 

\begin{assump}{Regular treatment assignment mechanism (Ignorability and Positivity):}
\label{assumption:regular}
The treatment assignment is conditionally independent of potential long-term outcomes (Ignorability) and all units have positive probability of being assigned to each action (Positivity) in the experimental dataset.
    \begin{gather}
    A_i \indep (Y_i(a), S_i(a))|X_i \ \forall a \in \mathbb{A}, i \in E \\
    0 < \pi(a|x) <1 \ \forall a \in \mathbb{A}, x \in \mathbb{X}
    \end{gather}
\end{assump}
Assumption 1 is satisfied when we have indeed conducted a randomized experiment, even if the probability of assignment to actions is conditional on observed covariates, as in our application.

\begin{assump}{Surrogacy:}
\label{assumption:surrogacy}
The treatment assignment is independent of long-term outcomes conditional on the surrogates in the experimental dataset.
    \begin{equation}
         A_i \indep Y_i\ |\ S_i,X_i,i \in E
    \end{equation}
\end{assump}

\begin{figure}[tb]

\begin{center}
\vspace{.3cm}
\scalebox{1.3}{
\begin{tikzpicture}[font=\sffamily, scale=1.2]
\begin{scope}[every node/.style={circle,thick,draw}]
    \node (A) at (0, 0) {A};
    \node (S) at (2, 0) {S};
    \node (Y) at (4, 0) {Y};
    \node (U) at (3, 1) {U};
    \node (X) at (3, 2) {X};
\end{scope}

\begin{scope}[>={Stealth[black]},
              every edge/.style={draw=black,thick}
              ]
    \path [->] (A) edge (S);
    \path [->] (S) edge (Y);
    \path[dotted] [->] (U) edge (S);
    \path[dotted] [->] (U) edge (Y);
    \path [->] (X) edge (Y);
    \path [->] (X) edge (A);
    \path [->] (X) edge (S);
    \path[dashed,bend right=30]
       [->] (A) edge (Y);  	
\end{scope}
\end{tikzpicture}
}
\caption{Directed acyclic graph representing causal relationships relevant to satisfying  the assumptions.}
\label{fig:dag}
\end{center}

\subcaption*{\scriptsize \textit{Note:} $A$ is the treatment, which is randomized (possibly conditional on $X$); $S$ are the surrogates; $Y$ is the long-term outcome, $U$ and $X$ are unobserved and observed covariates, respectively. This graph satisfies the Ignorability component of Assumption \ref{assumption:regular}. One way to satisfy Assumption \ref{assumption:surrogacy} is the absence of causal pathways from $A$ to $Y$ that do not go through $S$, that is, that the dashed edge is absent. One threat to the validity of Assumption \ref{assumption:comparability} is if an unobserved time-varying variable $U$ causes $S$ and/or $Y$ (dotted edges), so the observable relationship between $Y$ and $S$ is changing over time due to $U$.}
\end{figure}
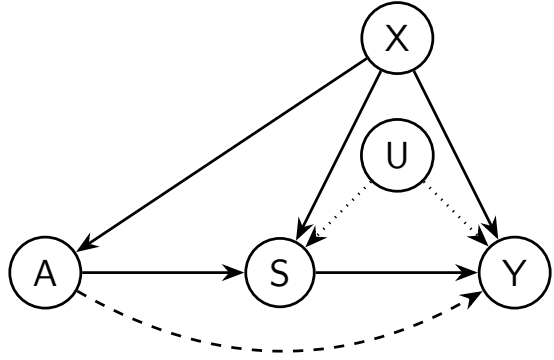

\noindent While there can be other ways to satisfy this assumption, surrogacy is perhaps most intuitively implied by a generative model in which the set of surrogates fully mediate the causal effects from treatment to the long-term outcome \citep[cf.][]{lauritzen2004discussion}, as depicted in Figure \ref{fig:dag} if the $A$ to $Y$ edge is absent.
In our empirical context, it means the effects of price discounts on long-term retention and revenue should occur via some intermediate outcomes we observe, e.g., content consumption and short-term revenue.
While it may have some testable implications, Assumption 2 is not directly testable.\footnote{This can also be described as an exclusion restriction, as in instrumental variables. Like that case this assumption has both testable and untestable implications. It might be tempting to regress the outcome on surrogate and treatment and test if the coefficient of treatment is zero. This naive test is not valid when there are unobserved confounders for the surrogate and outcome: conditioning on the surrogate or a ``collider'' in such a case will generate spurious correlation between treatment and confounder, and hence between treatment and outcome. See \cite{joffe2009related} for a more detailed discussion.}
Surrogacy is more plausible if we have a rich set of surrogates; perhaps this is more widely available given the increasing digitization of, e.g, commerce and media consumption (as in our application).

\begin{assump}{Comparability:}
\label{assumption:comparability}
The distribution of the long-term outcome conditional on the covariates and surrogates is the same across the experimental and historical datasets.
    \begin{equation}
           Y_i\ |\ S_i,X_i,i \in E \sim Y_i\ |\ S_i,X_i,i \in H
    \end{equation}
\end{assump}

\noindent In our case, this assumption implies that the distribution of long-term retention and revenue (conditional on content consumption and short-term retention and revenue) should be the same between the experimental and historical datasets.
Note that under comparability assumption we have:
   \begin{equation}
\tilde{Y}_i= \E_H[Y_i|S_i,X_i] = \E_E[Y_i|S_i,X_i]
\label{si}
\end{equation}
In other words, the conditional expectation of $Y_i$ in the experimental dataset is equal to the conditional expectation in the historical dataset, which is a quantity we can compute because in the historical dataset $Y_i$ is observed.
%\noindent To make Assumption 3 more plausible, we use the most recent historical data to do the estimation; that is, for the experiment run in 2018 we used the observed revenue data from 2015--2018 to estimate the 3-year revenue for subscribers in the experiment.\footnote{We can also directly test for this after the long-term outcomes in the experiment are realized, but not before.} 
This assumption would fail if the distribution of long-term outcome conditional on covariates and surrogates are changing between the experimental and historical datasets. For instance, if the intervention itself modifies the relationship between long-term outcome and surrogates, the two distributions will be different.
For example, in our empirical setting, it may be that, in the absence of an intervention, only very dedicated (i.e. high retention rate) subscribers read some categories of content; however, some actions might induce other, less dedicated subscribers to read that content.
For this reason, having similar (even unobserved) interventions in the historical data could strengthen our confidence in this assumption.
More extreme violations of this assumption can occur when measurement of a surrogate is changing (e.g., what counts as reading an article has a different definition in historical data). Note that, while not put in potential outcomes notion here or in \citet{athey2019surrogate}, one way for comparability to be satisfied involves observational causal inference about effects of $S$ on $Y$ using the historical data to succeed; thus, we expect that, as in observational causal inference, this is a very strong assumption that is often not exactly true. This motivates our consideration of weaker assumptions and the use of empirical evaluation in our application.

Given these assumptions, we prove that the surrogate index is valid for policy evaluation and optimization. Policy evaluation is the estimation of $V(\pi)$ for a given policy $\pi$. Policy optimization is finding a $\pi^*$ that maximizes $V(\pi)$. See Section \ref{policy} for more details about doing so in finite samples; here we simply consider the optimal policy defined on the population. We show that the value of a policy with respect to surrogate index is identical to its value on the long-term outcome; this in turn implies that the optimal policy with respect to the surrogate index coincides with that optimal policy with respect to long-term outcomes. We state the main results here and the proofs are in Appendix \ref{proof}. Let $\tilde{V}(\pi)$ denote the value of $\pi$ with respect to $\tilde{Y}$ rather than $Y$.  

\begin{prop}{Under Assumption 1-3, policy evaluation conducted on surrogate index identifies the true policy value defined on long-term outcomes.}
    \begin{equation}
           \tilde{V}(\pi) = V(\pi) \; \forall \pi \in \Pi
    \end{equation}
    \label{prop:surrogate_value}
\end{prop}
\noindent
Then, since the function being maximized is identical at all points, it is also identical at its maximum.

\begin{prop}{Under Assumption 1-3, policy optimization conducted on surrogate index recovers the true optimal policy.}
    \begin{equation}
           \underset{\pi \in \Pi}{{\mathrm{argmax}}} \ \tilde{V}(\pi) = \underset{\pi \in \Pi}{{\mathrm{argmax}}} \ V(\pi)
    \end{equation}
    \label{prop:surrogate_optimal}
\end{prop}

Proposition \ref{prop:surrogate_value} and \ref{prop:surrogate_optimal} are analytical results that could justify the approach developed here and employed in our empirical application. However, somewhat weaker assumptions than have been used for results for estimation of the ATE or CATEs are in fact sufficient for Proposition \ref{prop:surrogate_optimal}. 

Define real and surrogate-index-imputed CATEs, $\CATE_{aa^{\prime}}(x) = \E_E[Y(a)-Y(a^{\prime}) | X = x]$ and $\tCATE_{aa^{\prime}}(x) = \E_E[\tilde{Y}(a)-\tilde{Y}(a^{\prime}) | X = x]$. When, e.g., Assumption \ref{assumption:surrogacy} is violated (perhaps the set of surrogates does not fully mediate the treatment effect on long-term outcomes), the CATE estimated using surrogate index can be  biased (even with infinite data). That is, $\CATE_{aa^{\prime}}(x) \neq \tCATE_{aa^{\prime}}(x)$ for some $x \in \mathbb{X}$.
Here our aim is not estimating CATEs, but simply optimizing the policy. Bias in  CATEs (i.e. non-zero $\CATE_{aa^{\prime}}(x)-\tCATE_{aa^{\prime}}(x)$) does not result in a loss in the value of the optimized policy unless the bias changes the sign of that CATE.\footnote{Concern with getting the sign of the treatment effect correct using surrogates has featured prominently in the literature on the ``surrogate paradox'', in which various surrogacy definitions are satisfied by the effect on the surrogate and outcome have opposite signs; see, e.g., \citet{chen2007criteria,vanderweele2013surrogate,jiang2016principal}.}

Thus, we can introduce a somewhat weaker assumption, replacing Assumptions \ref{assumption:surrogacy} and \ref{assumption:comparability}, that is sufficient for policy optimization. The intuition that sign preservation is sufficient is that, for policy optimization purposes, we only care about identifying which is the best action for each unit, not how much better it is (i.e., we just need to correctly order the actions with respect to treatment effects, the magnitude of differences between actions do not matter).

\begin{assump}{Sign Preservation:}
The sign of conditional average treatment effects is the same for the surrogate index and the long-term outcome.
    \begin{equation}
         \sgn(\tCATE_{aa^{\prime}}(x)) = \sgn(\CATE_{aa^{\prime}}(x)) \; \forall a, a^{\prime}\in \mathbb{A}, x \in \mathbb{X}
    \end{equation}
    \label{assumption:sign}
\end{assump}
This is an assumption directly on CATEs, and so is not as readily interpretable with respect to the data-generating process. Nonetheless, we can reason about how this assumption may be more plausible in some settings than others. For example, in cases with a binary treatment, if we hypothesize that a treatment ``works'' (i.e., has a large positive effect) on some groups but not others, and this treatment has some small cost (which is incorporated into the definition of $Y$), then the distribution of CATEs may be bi-modal with no density near zero. This could contrast with other cases where theory might lead us to expect highly heterogeneous benefits and costs of the treatment (both incorporated into the definition of $Y$). For example, in our empirical application, for subscribers whose behavior is unaffected by a discount, this will reduce long-term revenue to varying degrees depending on how long they are retained; similarly, for those affected, this may affect long-run revenue in complex, heterogeneous ways. This highlights the value of empirical validation of surrogate-index-based policy optimization in our setting (Section \ref{validation}). Even in the favorable case where the distribution of CATEs is bi-modal with no density near zero, analysis with an impoverished set of covariates may result in loss. Say these available covariates are less informative about treatment effects; then the distribution of CATEs might have substantial mass near zero, raising the concern that any bias in CATE estimation may translate to selecting a sub-optimal policy when using a surrogate index.

One can analytically characterize the loss in policy optimization, much as \citet{athey2019surrogate} develop bounds on the bias for the ATE. Here we state this result, with details in Appendix \ref{proof}.

\begin{prop}
{There is a loss in the value of optimal policy only when the optimal action estimated on surrogate index is different than the true optimal action. The total loss, or regret, is:}

\begin{equation}
    \int_X {\tau}_{a^*\tilde{a}^*}(X) \cdot \mathbbm{1}_{\{a^*(X) \not = \tilde{a}^*(X)\}} \ dF(X) 
    %\{ \sgn(\tCATE(X)) \not = \sgn(\CATE(X))\}
\end{equation}
% \begin{equation}
%     \int_X |{\tau}_{a^*\tilde{a}^*}(X)| \cdot \mathbbm{1}_{\{a^*(X) \not = \tilde{a}^*(X)\}} \ dF(X) 
%     %\{ \sgn(\tCATE(X)) \not = \sgn(\CATE(X))\}
% \end{equation}
\begin{equation}
    a^*(X) := \{a \in \mathbb{A} | \CATE_{aa^{\prime}}(X) > 0 \ \forall a^{\prime} \in \mathbb{A} \}
\end{equation}
\begin{equation}
        \tilde{a}^*(X) := \{a \in \mathbb{A} | \tCATE_{aa^{\prime}}(X) > 0 \ \forall a^{\prime} \in \mathbb{A} \}
\end{equation}
\end{prop}

In summary, assumptions introduced in the surrogacy literature can be used to justify policy evaluation and optimization with a surrogate index. Furthermore, it is possible to relax these assumptions for policy optimization precisely because the optimal policy is only sensitive to the sign of treatment effects.

\subsection{Evaluating, Learning, and Implementing Targeting Policies}\label{policy}

We describe the off-policy evaluation and learning framework using the imputed long-term outcome $\tilde{Y}$ obtained via the procedure in Section \ref{surrogate index}.\footnote{In an abuse of notation, we now use $\tilde{Y}$ (rather than, e.g., $\hat{\tilde{Y}}$) to denote the actually imputed long-term outcome, which is estimated, while in Definition \ref{def:si} it denoted the true conditional expectation, as otherwise this makes some further expressions cumbersome.
}
Under assumptions articulated in the previous section, this can identify the same optimal policy as using the true long-term outcome $Y$. We use $\sim$ on variables or functions with parameters constructed with $\tilde{Y}$. We will describe each term generically but also make some connections to the quantities in our experiments. Readers familiar with counterfactual policy evaluation and learning may choose to skip to Section 
\ref{result} where we discuss the experiments and results.
%\ref{exploration}

\subsubsection{Off-policy Evaluation}
\label{eval}

In off-policy evaluation we use data collected under the design (or behavior) policy\footnote{In the reinforcement learning literature~\citep[e.g.,][]{sutton2018reinforcement}, the policy used to collect training data is called a behavior policy. We call it a design policy in our experimental setting.} $\pi_D$ to estimate the value of a counterfactual policy $\pi_P$. One popular choice of estimator is based on inverse probability weighting (IPW). The H\'ajek estimator, a normalized version of the Horvitz--Thompson estimator \citep{horvitz1952generalization}, is typically used to implement IPW \citep{sarndal1992model}. The H\'ajek estimate of the average outcome under an arbitrary targeting policy $\pi_P$ using data collected under a design or behavior policy $\pi_D$ is:
\begin{equation}
{\hat{\tilde{V}}}_{\text{IPW}}(\pi_P)=   \left(\sum_i\frac{\pi_P(A_i|X_i)}{\pi_D(A_i|X_i)}\right)^{-1}\cdot\sum_i \frac{\pi_P(A_i|X_i)}{\pi_D(A_i|X_i)} \cdot \tilde{Y}_i 
\label{ipsestimator}
\end{equation}
where $\tilde{Y}_i$ is the imputed outcome (e.g., predicted 3-year revenue), $A_i \in \{0,1,2,...,K-1\}$ is the action (e.g., discount) received by unit $i$ assigned by the design policy $\pi_D$, and $\pi_P$ is the probability of assigning unit $i$ to a given condition under the counterfactual policy that we want to evaluate.\footnote{The corresponding unnormalized Horvitz--Thompson estimator is:
$\frac{1}{n}\sum_i \frac{\pi_P(A_i|X_i)}{\pi_D(A_i|X_i)} \cdot \tilde{Y}_i$}
We will use $A_i = 0$ to denote the control and $A_i = 1$ to denote the treatment when actions are binary.\footnote{For example, when $A_i =1$ it means unit $i$ was in treatment and she was assigned to treatment with probability $\pi_{D}(1|X_i)$, and $\pi_P(1|X_i)$ is the probability that $i$ receives treatment under counterfactual policy $\pi_P$. Similarly, when $A_i=0$ it means unit $i$ was in the control and she was assigned to control with probability $\pi_{D}(0|X_i)$, and $\pi_{P}(0|X_i)$ is the probability that $i$ will be in control (or \textit{not} be treated) under counterfactual policy $\pi_P$.} The first term in Equation~\ref{ipsestimator} is simply a normalization term; the ratio between $\pi_P$ and $\pi_D$ is also known as the importance weight.
As specified by Assumption \ref{assumption:regular}, we need $\pi_{D}$ to be strictly positive for all unit--action pairs. Note that we do not require the policy being evaluated $\pi_{P}$ to have this property; it can be a deterministic policy. The Horvitz--Thompson estimator is unbiased but typically has higher variance. The H\'ajek estimator is biased in finite samples but consistent, and it typically has lower variance; it is therefore more widely used in practice.\footnote{For more discussion about of normalization in IPW estimation, see \citet[ch. 9]{owen2013monte} and \citet{khan2021adaptive}.} The main advantage of IPW is that it is fully non-parametric when the propensity scores are known and it does not require us to specify a model for the outcome process.

However, the IPW estimator has two main limitations: First, the H\'ajek estimator can still suffer from high variance. Second, when evaluating a deterministic policy $\pi_P$, it only uses observations for which the actions prescribed by the target policy $\pi_P$ and design policy $\pi_D$ agree (when they don't agree $\pi_P (A_i|X_i)$ is always zero). This reduces the effective sample size, especially when $\pi_P$ and $\pi_D$ are very different.\footnote{Two policies are similar if they tend to prescribe the same action for a given unit profile, the more often they prescribe different actions for a given unit, the more different they are.} Following \citet{robins1994estimation}, one way to improve upon IPW is by augmenting it with an outcome model $\mu$ to use all observations and further stabilize the estimator. This is known as the augmented inverse probability weighted (AIPW) or doubly-robust (DR) estimator \citep{dudik2014doubly}. Under the DR approach, the value of a policy $\pi_P$ can be estimated as:
\begin{equation}
{\hat{\tilde{V}}}_{\text{DR}}(\pi_P)  = \frac{1}{n}\sum_i \left ( \hat{\tilde{\mu}}(X_i, \pi_P) + \frac{\pi_P(A_i|X_i)}{\pi_D(A_i|X_i)} \cdot (\tilde{Y}_i - \hat{\tilde{\mu}}(X_i, A_i))\right),
\label{drestimator}
\end{equation}
where 
\begin{equation}
\hat{\tilde{\mu}}(X_i, \pi_P)  = \sum_{a \in \mathbb{A}} \pi_P (a|X_i) \cdot \hat{\tilde{\mu}}(X_i, a).
\end{equation}
The first term in Equation~\ref{drestimator}, $\hat{\tilde{\mu}}(X_i, \pi_P)$, is an outcome model that estimates the expectation of the imputed outcome for a random covariates profile $X_i$ and distribution of actions given by a policy $\pi$ using data from the experiment.
(In the most common case of evaluating a deterministic policy, $\hat{\tilde{\mu}}(X_i, \pi_P)$ is just $\hat{\tilde{\mu}}(X_i, a)$ for the action to which $\pi_P$ assigns units with covariate profile $X_i$.)
For example, in our empirical application, it corresponds to the estimated 3-year revenue for a subscriber profile $X_i$ under a particular discount $a$. Note that this outcome model $\hat{\tilde{\mu}}$ is different from the one for $\tilde{Y}$ in Equation \ref{si}; there the outcome is estimated as a function of surrogates and covariates using historical data, whereas $\hat{\tilde{\mu}}$ estimates outcome as a function of actions and covariates using the experimental data. The second term is the importance weight multiplied by the prediction error; it corrects the first term towards the direction of the long-term outcome by an amount that is proportional to the prediction error. For a deterministic target policy $\pi_P$ it does so whenever the actions prescribed by $\pi_D$ and $\pi_P$ agree. Note that the high variance of IPW estimators is from the importance weights (dividing by a small probability when $\pi_D$ is very unbalanced), this term vanishes if the prediction error is small. Both IPW and DR estimators are consistent, but DR estimation can achieve semiparametric efficiency \citep[see, e.g.,][]{robins1994estimation,hahn1998role,farrell2015robust}, and typically has lower variance. We use the DR estimator for policy evaluation. 

\subsubsection{Off-policy Optimization} \label{optimal targeting policy}
As shown in the previous section, policy optimization builds on CATE estimation. We focus on using doubly-robust estimation.\footnote{Estimation of CATE can also be implemented in different ways. \citet{hitsch2018heterogeneous} distinguish between what they label ``indirect'' approaches (which first estimate the outcome model as a function of covariates and actions and then take the difference between actions as treatment effects) and ``direct'' methods estimate the CATE directly without first estimating an outcome function (e.g., causal trees \citep{athey2016recursive}), causal forest \citep{wager2017estimation} and causal kNN \citep{hitsch2018heterogeneous}). This typology may be confusing to readers familiar with contextual bandit and policy learning literatures where, at least since \cite{dudik2014doubly}, ``direct methods'' are those using outcome regressions without IPW (i.e. what  \citet{hitsch2018heterogeneous} label ``indirect'').}
We can first construct a doubly-robust score for each unit--action pair (which also has the interpretation of an estimate of an individual potential outcome) \citep{robins1994estimation, chernozhukov2016locally, dudik2014doubly, athey2017efficient, zhou2018offline}:
\begin{equation}
\hat{\tilde{\gamma}}_a(X_i) = \hat{\tilde{\mu}}(X_i, a) + \frac{\tilde{Y}_i - \hat{\tilde{\mu}}(X_i,a)}{\pi_D(a|X_i)}\cdot 1_{\{A_i=a\}}  \label{inf-function}.
\end{equation}
These doubly-robust scores are equal to the prediction of an outcome model $\hat{\tilde{\mu}}(X_i,a)$ plus a correction term based on IPW; the correction is applied if and only if the action being evaluated is the same as the action taken. This is intuitive because the correction term depends on $\tilde{Y}_i$, which is the outcome under a realized action $A_i$; it is informative only when the action being evaluated is the same as $a$, otherwise the term drops out and the doubly-robust scores reduce to the outcome model. The CATE, relative to the control, given a covariate profile $x$ can then be estimated as:
\begin{equation}
\hat{\tilde{\tau}}_a(x) = \frac{1}{n} \sum_{i: X_i = x} \left( \hat{\tilde{\gamma}}_a(X_i) - \hat{\tilde{\gamma}}_0(X_i) \right).
\label{tau-a}    
\end{equation}
We can use these doubly-robust scores for policy optimization \citep{murphy2001marginal,dudik2014doubly} by solving a cost-sensitive classification problem.\footnote{When $\pi_D(x)$ must be estimated, this approach comes with guarantees on asymptotic regret compared with the true optimal policy \citep{athey2017efficient,zhou2018offline}.}
That is, the estimated optimal policy is:
\begin{equation}
\hat{\tilde{\pi}}^* = \text{argmax}_{\pi \in \Pi} \ \ \frac{1}{n} \sum_i \ \left(\hat{\tilde{\gamma}}_1(X_i)-\hat{\tilde{\gamma}}_0(X_i) \right) \cdot \left( 2\pi(X_i)-1 \right), 
\label{policy-binary}    
\end{equation}
or, in multi-action case:
\begin{equation}
\hat{\tilde{\pi}}^* = \text{argmax}_{\pi \in \Pi} \ \ \frac{1}{n} \sum_i <\hat{\tilde{\gamma}}(X_i),\pi(X_i)>,
\label{policy-multi}    
\end{equation}
where $\hat{\tilde{\gamma}}(X_i) = (\hat{\tilde{\gamma}}_0(X_i), \hat{\tilde{\gamma}}_1(X_i),...,\hat{\tilde{\gamma}}_k(X_i))$ is a vector of doubly-robust scores based on Equation \ref{inf-function} and $\pi(X_i)$ is a vector of probabilities with which the policy assigns a unit to each action. $<\cdot>$ is the dot product between 
vector valued $\hat{\tilde{\gamma}}(X_i)$ and $\pi(X_i)$. 

In the cost-sensitive classification problem, for each unit, the correct label is the action that corresponds to the highest doubly-robust score, and the loss for classifying a unit to action a, when the correct label is $a^*$, is $\hat{\tilde{\gamma}}_{a^*}(X_i) - \hat{\tilde{\gamma}}_a(X_i)$, which is the loss the imputed outcome (e.g., predicted 3-year revenue) when a unit is assigned to a suboptimal action. In multi-action cases, a cost-sensitive binary classification is done on every pair of actions, and the final action is chosen by a majority vote. In practice, the policy class $\Pi$ is often restricted by the choice of a specific type of classifier (e.g., logistic regression or decision trees for interpretation or transparency reasons), or by using only a subset of covariates in the classifier (while still using all information to construct the doubly-robust scores). A practical advantage of this approach is that once the doubly-robust scores or labels are constructed, we can plug them into off the shelf classifiers to optimize the policy.

\subsubsection{Policy Implementation and Exploration} \label{policy implementation}
While we have estimated the optimal policy, it is typically desirable to account for remaining statistical uncertainty and continue randomized exploration, which can be particularly important if there is non-stationarity, i.e., changes in the environment that make a policy that is optimal today no longer optimal in the future. While other approaches can be suitable, we find particularly suitable a variant of Thompson sampling, bootstrap Thompson sampling (BTS) \citep{eckles2014thompson,lu2017ensemble,osband2016deep}, that is readily implemented with models for which Thompson sampling might be cumbersome to implement; see \citet{eckles2019bootstrap} and \citet{osband2017deep} for reviews. 
We use BTS as a heuristic approach to adding randomized uncertainty-based exploration to the estimated optimal targeting policy where a unit $i$ is assigned to action $a$ with probability proportional to the fraction of times an action is estimated to be optimal across all bootstrap replicates of the data. That is, 
\begin{equation}
\hat{\tilde{\pi}}_{\text{BTS}}(a|X_i) = \frac{1}{R} \sum_{r = 1}^R 1_{\{\hat{\tilde{\pi}}^*_r(X_i) = a\}},
\label{bts}    
\end{equation}
where $\hat{\tilde{\pi}}^*_r$ is the policy estimated according to Equation \ref{policy-binary} or \ref{policy-multi} on the $r$th bootstrap replicate.\footnote{In cases where a unit is always or never assigned to some conditions, we may want  to impose a probability floor and ceiling to ensure that all units have positive probability being assigned to all conditions, thereby satisfying Assumption \label{assumption:regular}.}
%\newpage

\subsection{Summary of the Methods} \label{method summary}
We summarize the key steps in combining these methods as follows:

0) Identify the long-term outcome of interest ($Y$), intervention ($A$), covariates ($X$), and surrogates ($S$). 

1) Run a randomized experiment through a design policy $\pi_D$ to generate experimental data $(X,A,S)$. Gather historical data $(X,S,Y)$.

2) Impute the missing long-term outcomes in the experiment using the surrogate index $\tilde{Y}$ through Equation \ref{si}.

3) Do policy optimization using imputed long-term outcomes $\tilde{Y}$ to get an estimated optimal policy $\hat{\tilde{\pi}}^*$ through Equation \ref{inf-function} and \ref{policy-binary} or \ref{policy-multi}.

4) Implement the estimated optimal policy $\hat{\tilde{\pi}}^*$, potentially with added randomization as in $\hat{\tilde{\pi}}_{\text{BTS}}$ through Equation \ref{bts}.

5) Consider Step 4 as running a new randomized experiment with $\hat{\tilde{\pi}}_{\text{BTS}}$ being the new $\pi_D$, and repeat Step 1 -- 4 as desired.

\begin{figure}
    \centering
    \includegraphics[width=0.85\textwidth]{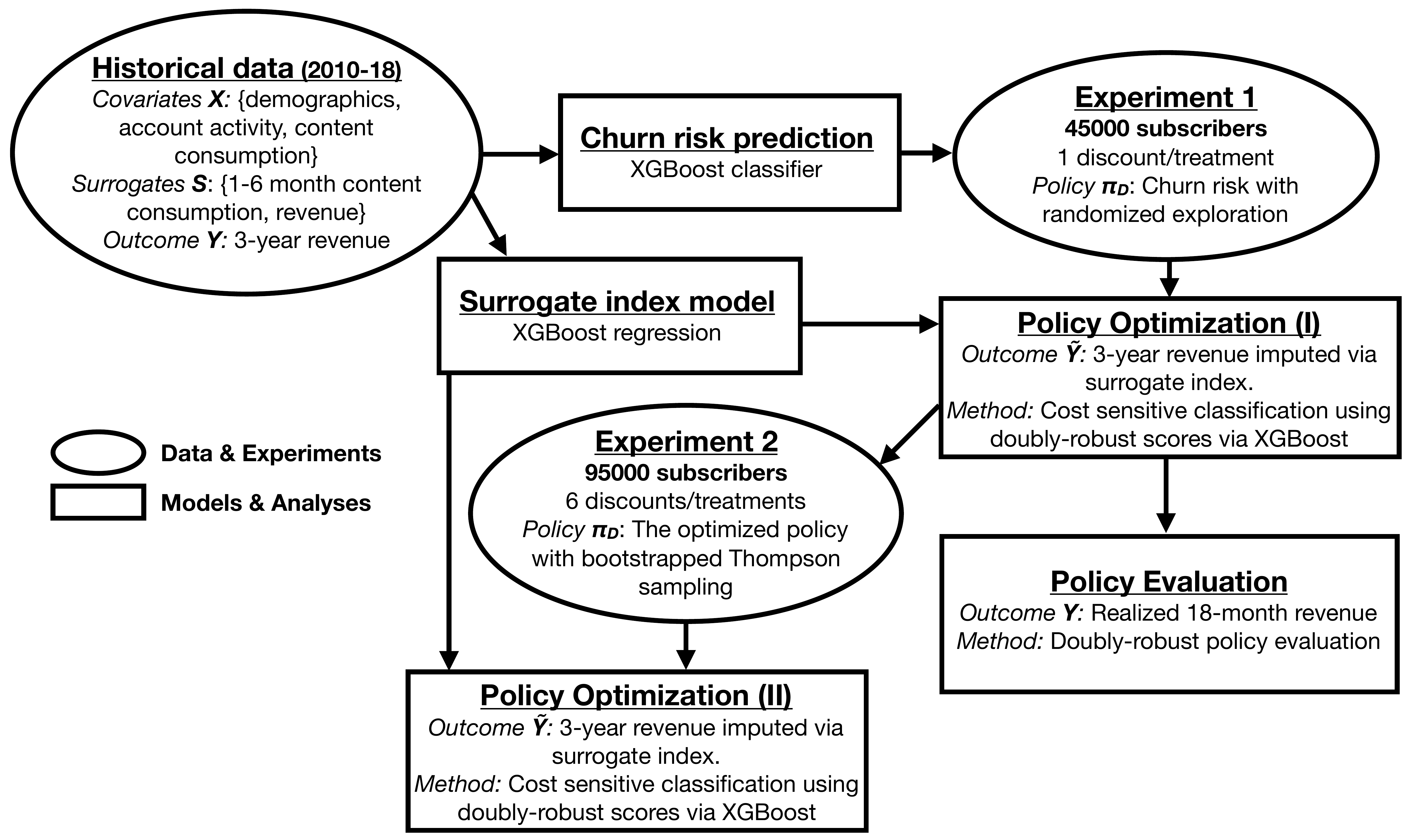}
    
    \caption{Summary of observational and experimental data and analyses.}
    \subcaption*{\scriptsize \textit{Note:} Historical observational data is used to train a churn prediction model and a model for long-term outcomes (producing a surrogate index). Experiment 1 uses the churn predictions in randomly assigning subscribers to treatments. Using the data from Experiment 1, we learn a policy using the surrogate index, which is then used in (a) the design of Experiment 2 and (b) evaluations compared with actual 18-month revenue. Similarly, we learn a policy from the Experiment 2 data and the surrogate index.}
    \label{fig:flowchart}
\end{figure}

\section{Experiments and Results} \label{result}

We now turn to applying and evaluating this approach in the context of reducing churn at \emph{The Boston Globe}, where we offer discounts to existing subscribers. Figure \ref{fig:flowchart} gives an overview of how the historical observational data and two field experiments relate to each other and the main analyses.
Experiment 1 randomized subscribers to receive a discount or not. We then optimized the targeting policy using results from Experiment 1 and a surrogate index constructed from historical data, the surrogates we use are: content consumption (number of articles read in each of the 20 most visited sections\footnote{The sections are: metro, sports, news, lifestyle, business, opinion, arts, Sunday magazine, ideas, search, member center, south, spotlight, page not found, nation, north, magazine, circulars, politics.} on \textit{The Boston Globe}'s website) and revenue over the first 6 months. We selected these surrogates based on the following reasoning. First, revenue captures whether a subscriber has already churned, as well as whether they have perhaps received other discounts (e.g., via reactive churn management). Second, subscribers get value from their subscription primarily by consuming articles and other content on \emph{The Boston Globe}'s website. We expected that some of this content is more differentiated from that otherwise available (e.g., local sports coverage). These surrogates could be measured over shorter or longer periods. Intuitively, the longer we wait, the better we can estimate the long-term revenue, but firms also want to learn the optimal policy quickly so we can implement it. In particular, we should expect that it will be important to observe revenue and consumption for some time after the discounts expire. Given these considerations, we used surrogates computed over 6 months of data.

We implemented the policy, with additional randomized exploration, in Experiment 2. Once 18 months had passed since the start of Experiment 1, we were able to compare the performance of the policy we learned using the surrogate index to that we would have learned using the longer-term, 18-month outcomes.\footnote{We use the most recent historical data to do the imputation; that is, for Experiment 1, run in 2018, we used the observed revenue data from 2015--2018 to estimate the 3-year revenue for subscribers in the experiment.}
All treatment effects are from intent-to-treat (ITT) analyses that do not condition on potentially endogenous post-treatment behaviors, such as opening the email or redeeming the benefit. 
We report the survival curves and treatment effects estimated from the resulting data in both experiments in Appendix \ref{treatment effects}. In this section, we focus on the experiment design, policy learning, and surrogate index validation results.

\subsection{Experiment 1} 
\label{first_cohort}

As is typical of a new effort in proactive churn management, we lacked prior experimental data in which subscribers were assigned to discounts. However, we anticipated that the discount treatment would not have substantial beneficial effects on subscribers with a low probability of churning. Thus, we assigned subscribers to treatment using a design policy $\pi_D$ in the first experiment that balances exploration and exploitation; we do so by assigning subscribers with higher predicted churn probability into treatment with higher probability, while ensuring that all subscribers $0 < \pi_D(X_i) < 1$, thus satisfying Assumption \ref{assumption:regular}; see Appendix \ref{churn prediction} and \ref{design policy} for a more detailed discussion. This assigned 806 subscribers to receive a discounted subscription rate (\$4.99 per week) for 8 weeks.  

We estimate the optimal policy via the binary cost-sensitive classification (Equation \ref{policy-binary}) on 
%both observed mid-term (18-month) revenue and 
imputed long-term revenue, defined as either 18-month or 3-year revenue. In this section, we focus on the policy using imputed 3-year revenue; we return to the policy using imputed 18-month revenue in our validation in Section \ref{validation}. We first construct doubly-robust scores for each subscriber using Equation \ref{inf-function} where $\hat{\tilde{\mu}}$ is estimated using XGBoost via cross-fitting.\footnote{Cross-fitting means that $\hat{\tilde{\mu}}$ for individual $i$ is estimated \textit{without} using $i$'s own data in the training process. We can split data randomly into $n$ folds, then $\hat{\mu}$ for individuals in a given fold is trained only using data from the other $n-1$ folds, it reduces over-fitting and improves efficiency \citep{athey2017efficient, zhou2018offline}. We use $n=3$ in our estimation.} We then split the data into training (80\%) and test sets (20\%) and use XGBoost as the classifier with hyper-parameters tuned via cross-validation. The policy learned using the surrogate index, $\hat{\tilde{\pi}}$,  would treat 21\% of subscribers in the experimental data. We evaluate policy performance on the test data using the doubly-robust estimator as in Equation \ref{drestimator}. According to the surrogate index, it would generate a \$40 revenue increase per subscriber (95\% confidence interval [\$10, \$75]) over 3 years compared to the current policy that treats no one, which is \$1.7 million dollars in total for subscribers in the first experiment.

We use tools in interpretable machine learning to look at what variables are most important in determining the optimal policy, and how the optimal policy depends on these variables (see Appendix \ref{interpretation}). The top three variables are risk score (predicted risk of churn), tenure, and number of sports articles read in the last 6 months. The optimized policy treats subscribers with shorter tenure (more recently registered subscribers) at a higher rate. The relationship between number of sports articles read and treatment is not monotone: The fraction treated is low for very inactive and very active subscribers, but higher for subscribers in between. The relationship with risk score is interestingly also not monotone, for subscribers with the highest risk scores the treatment fractions are higher, this is consistent with our prior. But for some subscribers with very low risk scores, the treatment probabilities are even higher. This also highlights potential blind spots of targeting solely based on risk scores. 

\subsection{Experiment 2}
\label{sec:second_cohort}
Having learned a policy using the first experiment, we turned to exploiting this knowledge and further learning through experimentation in a second experiment. Furthermore, the success of the first experiment prompted creating and trying a larger set of 6 treatments: a thank you  email, a \$20  gift  card, a discount to \$5.99 for 8  weeks, a discount to \$5.99 for 4  weeks, a discount to \$4.99 for 8 weeks (the same as the intervention in the first experiment), and a discount to \$3.99 for 8 weeks.

We use the learned policy based on imputed 3-year revenue --- with two modifications --- to allocate subscribers to treatments.
First, as discussed in Section \ref{policy implementation}, adding randomization to an estimated optimal policy is a desirable practice especially in a potentially non-stationary environment. We added randomization to the optimized policy through bootstrap Thompson sampling as in Equation \ref{bts}. This assigned 5,688 subscribers to treatments. Second, since all but one of the treatments were new, the learned policy was not directly informative about which non-control actions to take; therefore, conditional on a subscriber being assigned to treatment, we assigned them to the 6 non-control conditions uniformly at random. For future subscribers, we can learn and implement an optimal policy over all interventions based on the results from Experiment 2. 

We optimize the policy via multi-class cost-sensitive classification (Equation \ref{policy-multi}) using data from Experiment 2 following a similar procedure as in Experiment 1. The optimized policy using the surrogate index, $\hat{\tilde{\pi}}$, allocates around a quarter of subscribers each to control, the thank you email, and the two smallest discounts; a few subscribers are allocated to other actions (Table \ref{tab:cohort2_optimal_actions}). This optimized policy improves 3-year revenue by \$30 per subscriber (95\% confidence interval [\$12, \$50]) relative to the status quo that treats no one, such that it would have generated \$2.8 million for subscribers in Experiment 2.

\begin{table}[htb] \centering 
  \caption{Distribution of optimal actions estimated from Experiment 2} 
  \label{tab:cohort2_optimal_actions} 
  \bigskip
  \def\arraystretch{1.3}
\begin{tabular}{ c r } 
\hline
Action & Percentage\\
\hline \hline
control &  23\% \\
\hline 
thank you email only & 25\% \\
\hline
gift card & <1\% \\
\hline
\$5.99/8 weeks & 25\% \\
\hline
\$5.99/4 weeks & 27\%\\
\hline
\$4.99/8 weeks &  <1\%\\
\hline
\$3.99/8 weeks &  <1\%\\
\hline
\end{tabular}
\subcaption*{\scriptsize \textit{Note:} \emph{Percentage} is the percentage of subscribers in Experiment 2 that are assigned to this action according to the policy optimized using the surrogate index, $\hat{\tilde{\pi}}$.}
\end{table}

We further compare the two experiments to see whether there are significant changes in the environment in terms of covariate and concept shift (Appendix \ref{non-stationarity}). When the environment is stationary, it is more efficient to pool data from the two experiments together to estimate the optimal policy for future subscribers, and when the environment is substantially changing, it is better to down-weight observations from the first experiment using a time-decaying case weight~\citep[e.g.,][]{russac2019weighted}. We only use data from the second experiment to estimate the optimal policy because there is some evidence for concept shift and there is only one common treatment condition between the two experiments.

\subsection{Surrogate Index Validation and Comparison} \label{validation}
The assumptions underlying surrogate-index-based policy learning are strong, and it is often implausible that they are strictly true; this is similar to, e.g., doubts about conditional ignorability in observational causal inference or the exclusion restriction in instrumental variables analyses. Thus, like in those settings \citep[e.g.,][]{dehejia2002propensity,gordon2019comparison, eckles2021bias}, it is often valuable to empirically evaluate the results of our approach when that is possible (i.e., when we \emph{do} observe long-term outcomes).
Researchers can wait until the true long-term outcomes are observed, and then compare the effect estimates and policies based on the surrogate index with those based on the true long-term outcomes. Here it takes 3 years to observe the long-term outcome \emph{The Boston Globe} is targeting for; instead, we use 18-month revenue (from August 2018 to February 2020), which is already realized at the time this is written, as the long-term outcome and repeat the analysis. Policy values are estimated using the doubly-robust approach as in Equation \ref{drestimator}, except that the outcomes we use are observed $Y_i$, not imputed $\tilde{Y}_i$.

We first look at how well the surrogate index recovers the treatment effect estimated on the true long-term outcome. We then evaluate it by looking at how it performs against a benchmark policy that is learned on some short-term proxies of the long-term outcomes (e.g., 1--6 month revenue) and a policy learned on the true long-term outcome (e.g., realized 18-month revenue). We also look at how the performance changes if we chose a different subset of surrogates. All policy values here are defined relative to the status quo of treating no one. We report confidence intervals from 1,000 bootstrap draws on the test data.

\begin{figure}
        \centering
        \begin{subfigure}{0.45\textwidth}
            \centering
          \includegraphics[width=\textwidth]{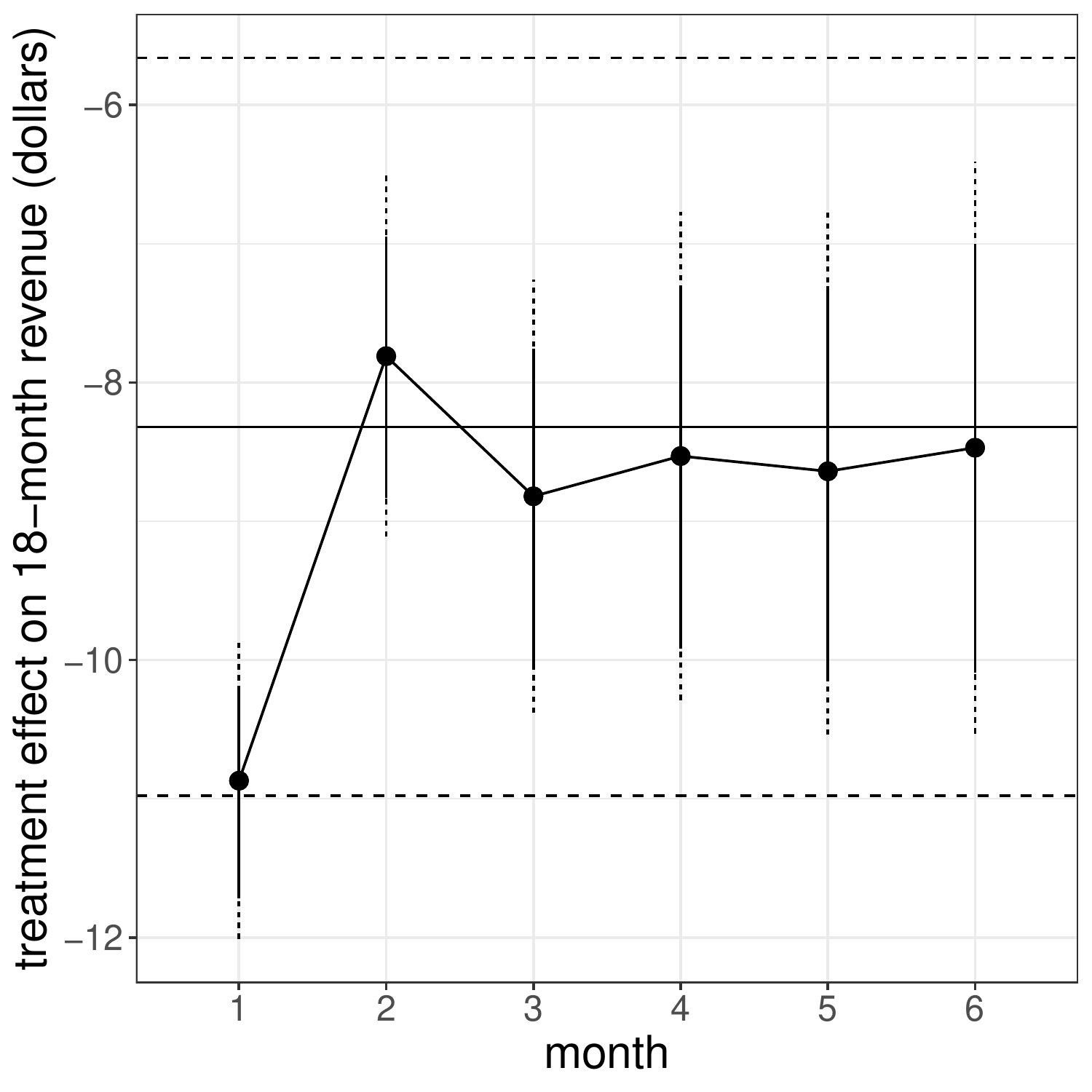}
            \caption
            {{\footnotesize (a) ATT on revenue using surrogate indices estimated with data from the first 1--6 months. The horizontal lines are the ATT estimated with true 18-month revenues and its 95\% confidence interval. The solid and dashed vertical lines are 75\% and 95\% confidence intervals, respectively.}}    
                     \label{fig-att}
        \end{subfigure}
        \hfill
         \begin{subfigure}{0.45\textwidth}   
            \centering 
         \includegraphics[width=\textwidth]{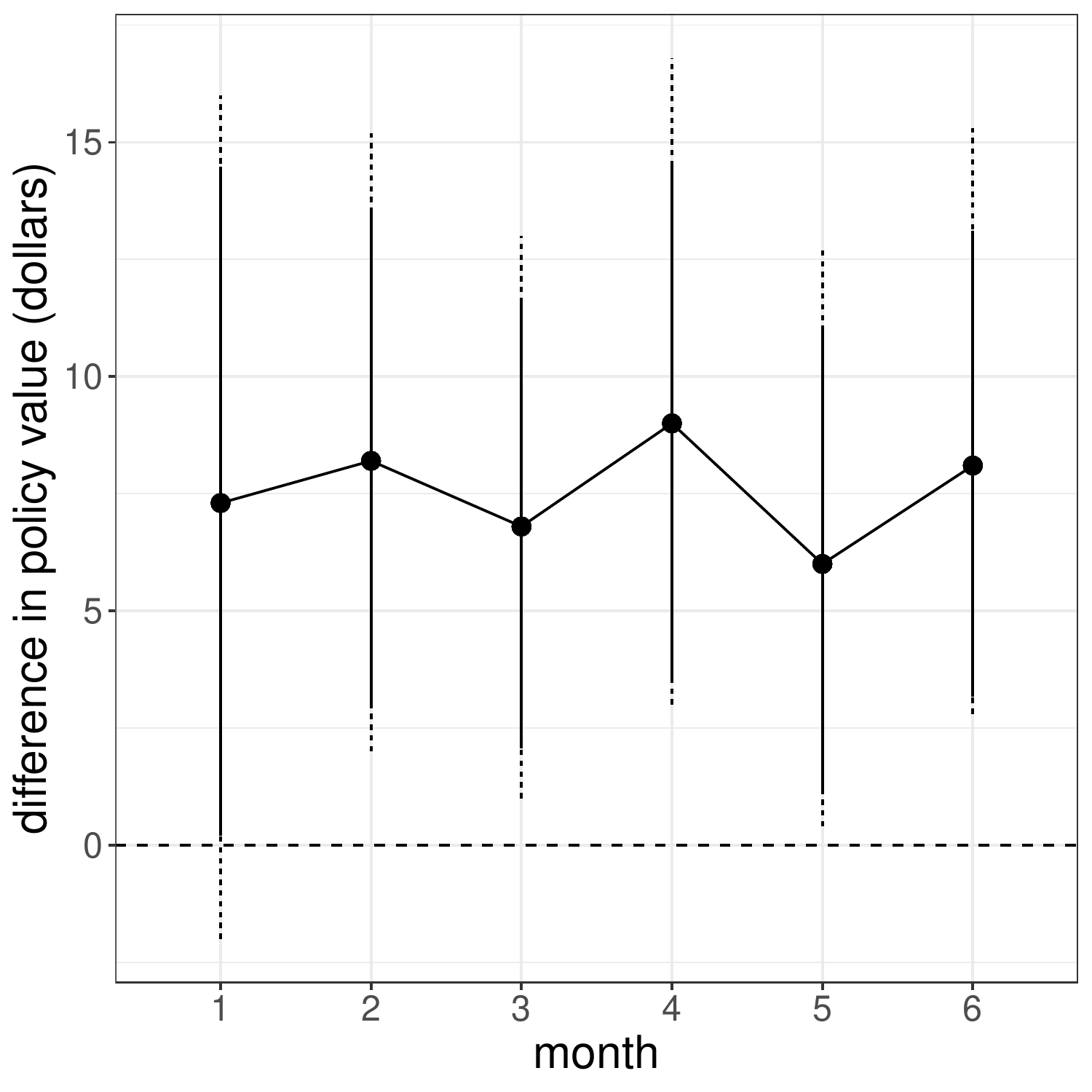}
            \caption[]%
            {{\footnotesize (b) The value difference between policies optimized on surrogate indices constructed with surrogates from the first 1--6 months and the current policy. Except for a single month, they outperform the status quo. The solid and dashed vertical lines are 75\% and 95\% confidence intervals, respectively.}} 
            \label{fig-si}
        \end{subfigure}
        \vskip\baselineskip
        \begin{subfigure}{0.45\textwidth}  
            \centering 
       \includegraphics[width=\textwidth]{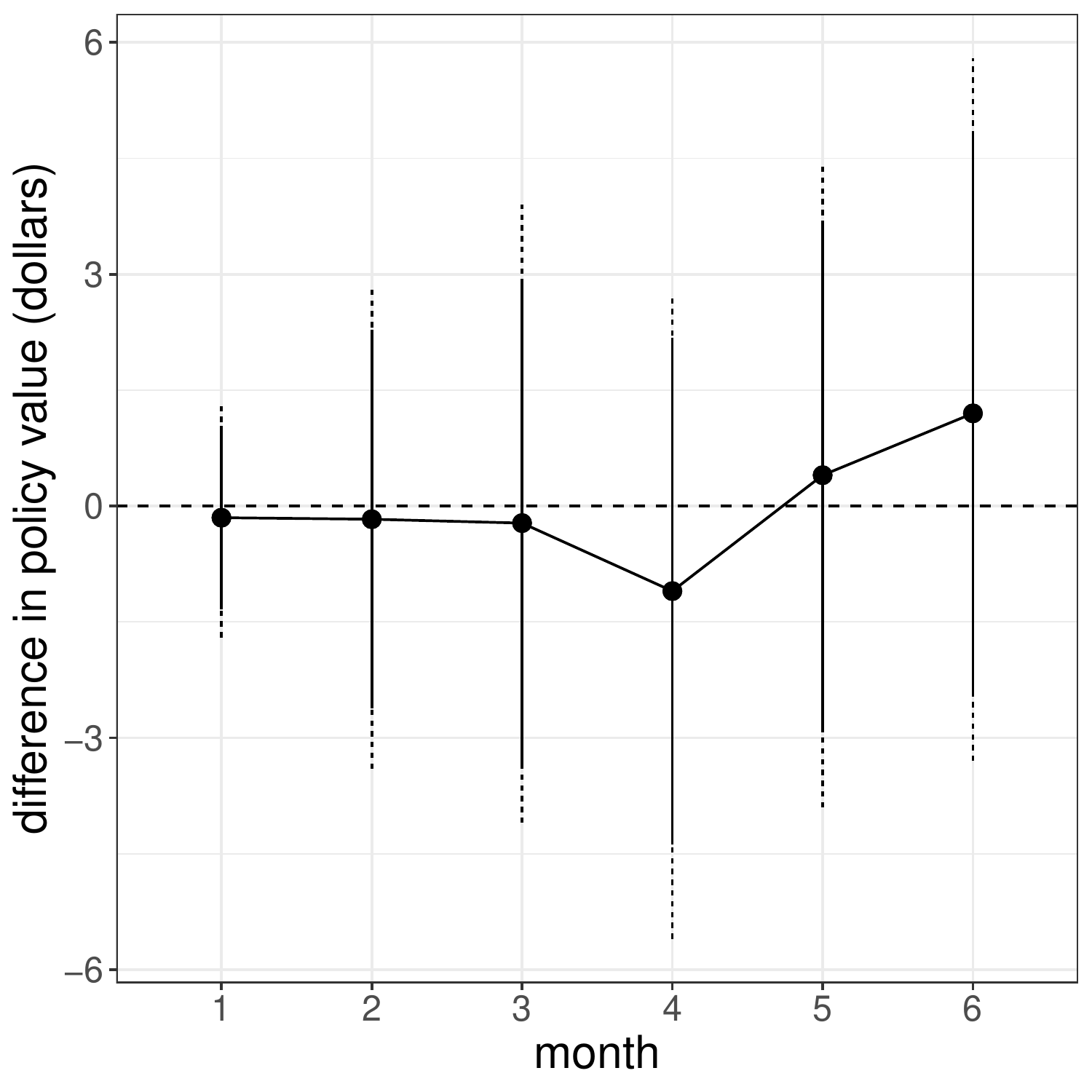}
            \caption[]%
            {{\footnotesize (c) The value difference between policies optimized with a single short-term proxy (revenues from the first 1--6 months) and the current policy. The value is indistinguishable from the status quo. The solid and dashed vertical lines are 75\% and 95\% confidence intervals, respectively.}}   
                    \label{fig-proxy}
        \end{subfigure}
                \hfill
        \begin{subfigure}{0.45\textwidth}   
            \centering 
          \includegraphics[width=\textwidth]{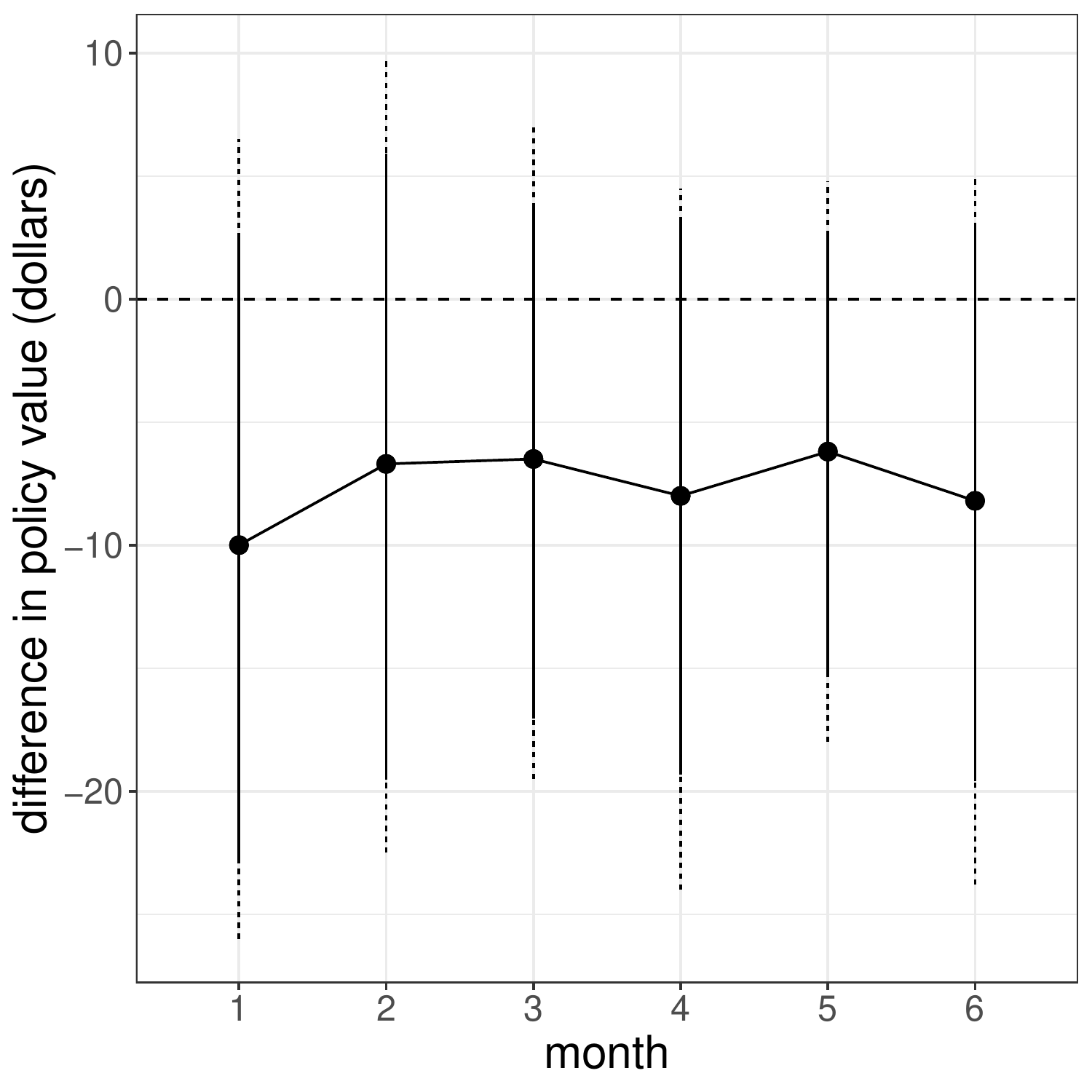}
            \caption[]%
            {{
            \footnotesize (d) The value difference between policies optimized on surrogate indices constructed with surrogates from the first 1--6 months and true outcomes. They are statistically indistinguishable. The solid and dashed vertical lines are 75\% and 95\% confidence intervals, respectively.}}
                    \label{fig-ground-truth}
        \end{subfigure}
        \caption[]
        { Empirical validation using Experiment 1 of using the surrogate index for treatment effect estimation and policy learning.} 
    \end{figure}

First, we look at how the average treatment effect on the treated (ATT) calculated using the surrogate index compares with ATT calculated using the true outcome (Figure \ref{fig-att}). After the first month, the surrogate-index-based ATT estimates are indistinguishable from the estimates using realized 18-month revenue. That the 1-month surrogate-index-based ATT is distinguishable from those using surrogates computed on longer periods may indicate that one month is too short a period; this is intuitively consistent as the treatment is a 8-week discount, so no reaction to the subsequent price increase is yet observed.
Note that the confidence intervals of ATT estimated on true outcomes are wider than the ones estimated on surrogate index. When the surrogacy assumption holds, it is more efficient to estimate the treatment effect on surrogate index because it discards irrelevant variation in the long-term outcome.

Next, we look at the value of surrogate-index-based policy (Figure \ref{fig-si}). All results are significantly better than the status quo except when we only use information from the first month; recall that the discount ends after 8 weeks. By contrast, optimizing the policy directly on short-term proxies (1--6 month revenue) does not detectably outperform the status quo (Figure \ref{fig-proxy}). We also compare the surrogate-index-based policy with a policy learned on the true long-term outcome (Figure \ref{fig-ground-truth}). Although all the point estimates of the value difference are negative, none of them is distinguishable from zero; the difference between the value of policy learned on surrogate indices using the first 6 month and true outcomes is -\$8 per subscriber (95\% confidence interval [-\$24, \$5]). This comparison does not take into account the gain in time and opportunity cost by implementing an optimized policy at 6 vs. 18 months. These two policies also agree on 72\% of subscribers, i.e., they assign them to the same treatment condition. This is encouraging, but it is also contributes to imprecision in estimating differences between them, as they estimates are determined by the long-term revenue of a smaller number of subscribers.

Lastly, we compare the performance of policies learned on surrogate indices constructed using only content consumption information, only short-term revenue, and both; the three approaches are not detectably different, though there is substantial uncertainty, so this does not rule out relevant differences (Appendix \ref{surrogate_choice}).\footnote{\cite{athey2019surrogate} suggests that when the surrogacy condition holds, the smallest set of surrogates has the highest precision in estimating the treatment effect.}

%\clearpage
%\newpage
\section{Conclusion} \label{conclusion}
Many applied problems, from the subscriber management problem studied here to others in business, medicine, public policy, and social sciences, where there is a need to personalize interventions to optimize some long-term outcomes, can be fruitfully characterized as learning a targeting policy for some long-term outcomes. 
Here we advance the practice of policy learning by incorporating the use of a learned surrogate index to impute the long-term outcomes. We first show analytically when a surrogate index is valid for policy evaluation and optimization in place of true unobserved long-term outcomes. Then to validate our approach empirically, we run two large-scale experiments that prescribe who should be targeted with what incentives in order to maximize long-term subscription revenue for \emph{The Boston Globe}. Combining data from the first experiment and the passage of time, we show that the policy optimized on long-term outcomes imputed by a surrogate index outperforms a policy optimized on a short-term proxy of the long-term outcomes and that it performs similarly to the policy optimized on true long-term outcomes. We then implement the optimized policy with additional randomized exploration so that we can respond to potential non-stationarity and update the optimized policy after each experiment. The total 3-year revenue impact of implementing policies optimized using the surrogate index, relative to the status quo, in the two experiments sums to \$4--5 million. Our paper adds to and complements a recent and growing literature in marketing on policy evaluation and learning \citep[e.g.,][]{hitsch2018heterogeneous,simester2019efficiently,simester2019targeting,yoganarasimhan2020design} and empirical work in proactive churn management \citep[e.g.,][]{ascarza2018retention} by focusing on optimizing targeting policies for long-term retention and revenue.

A natural question is how to choose surrogates when imputing long-term outcomes. If we have the generative model in Figure \ref{fig:dag} in mind, we want to choose variables that lie on the causal path from treatment to long-term outcomes, as suggested by domain knowledge or theory.  
We also want to choose surrogates that are observable shortly after the intervention so that the policy can be learned quickly. These two considerations may be in tension. 
If relevant experiments have been conducted in the past then the quality of surrogates can be evaluated on the realized long-term outcomes, as we have done here. 
Surrogates that are highly predictive of the outcome are potential candidates but there is no guarantee that they will produce high policy values, as predicting the outcome level is a different task than predicting the treatment effect or learning the policy. Future research may further examine selection of potential surrogates. In practice, we may only have noisy measurements of such surrogates; thus, a fruitful direction for future work may be incorporating recent developments from mediation analysis with multiple noisy measurements \citep{ghassami2021proximal}.  Finally, since surrogacy is fundamentally a question about the underlying causal mechanism, once some surrogates have been shown to be valid for a given problem, they may be likely to remain valid for similar problems in the future. For example, we showed short-term revenues and content consumption are suitable surrogates for the effect of price discounts on long-term retention and subscription revenues, so the firm can tentatively rely on this assumption as they continue to iterate on targeting policies. We can imagine building such a knowledge base for different sets of problems and long-term outcomes as more empirical researchers work in this general framework.

The present work is not without important limitations.
Some of these are limitations of the approach as developed here.
For example, it is directly applicable when there is essentially no constraint on how many units can be treated, as in our case. When there is a budget constraint and heterogeneous treatment costs, a policy can be optimized based on the ratio between individual-level treatment effect and the cost of treatment as in \cite{sun2021treatment}.
There are also important limitations to the strength of the conclusions from our empirical application. For example, while we were not able to detect differences in performance between the surrogate-index-based policy and one based on true long-term outcomes, this may reflect remaining statistical uncertainty in estimating this contrast; similar considerations apply to other comparisons, such as between the value of policies using different sets of surrogates. More generally, the quite promising results observed here may not be indicative of what practitioners can expect in other, even somewhat similar subscriber management settings, perhaps especially if a very different variety of actions are used. Thus, we hope that subsequent work offers both further methodological development and empirical validation.

\section*{Acknowledgements}
{Correspondence should be addressed to J.Y. (jeryang@hbs.edu) and D.E. (eckles@mit.edu). This research was supported in part by a grant from Boston Globe Media. We thank Boston Globe Media and particularly Jessica Bielkiewicz, Thomas Brown, Ryan McVeigh, and Shannon Rose for their partnership in conducting the field experiments. This work benefited from comments by Susan Athey, John Hauser, G\"unter Hitsch, Duncan Simester, participants in seminars at MIT, the Harvard Business School Digital Doctoral Workshop, the NeurIPS CausalML Workshop and the Quantitative Marketing and Economics (QME) Conference.}

\bibliography{ref.bib}{}
\bibliographystyle{plainnat}

\clearpage
\pagebreak
{
\section*{\center{\LARGE{\textmd{Online appendix for:\\ ``Targeting for long-term outcomes''}}}}
}
\vspace{1cm}

\begin{appendices}

%\ECHead{Supplemental Material}
\counterwithin{figure}{section}
\counterwithin{table}{section}
\section{New York Times Example}

\begin{figure}[!htb]
    \centering
    \includegraphics[scale=0.5]{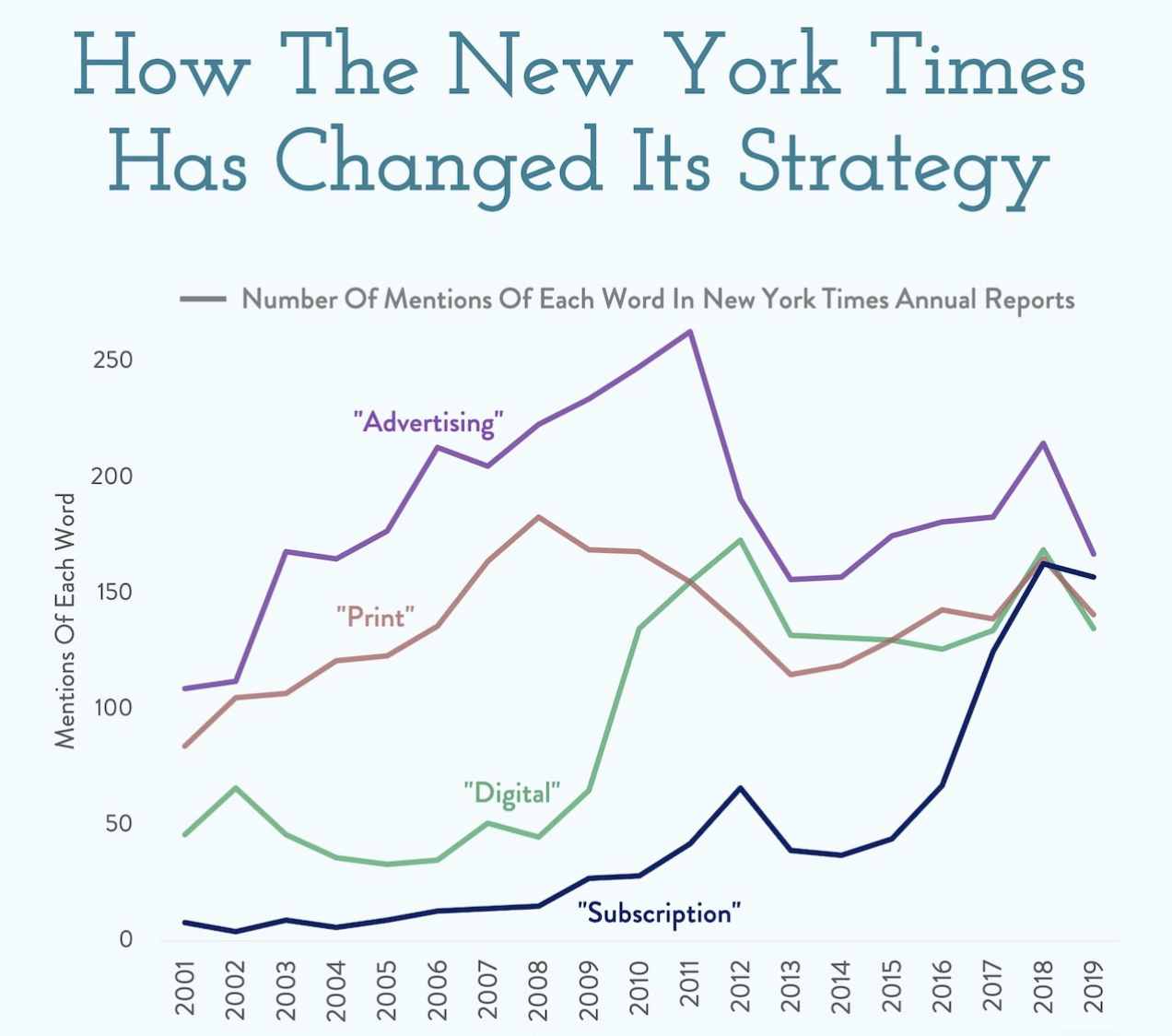}
    \bigskip
    \caption{Number of mentions of keywords in annual report over time (Source: chartr)}
    \label{nyt1}
\end{figure}

\begin{figure}[!htb]
    \centering
    \includegraphics[scale=0.6]{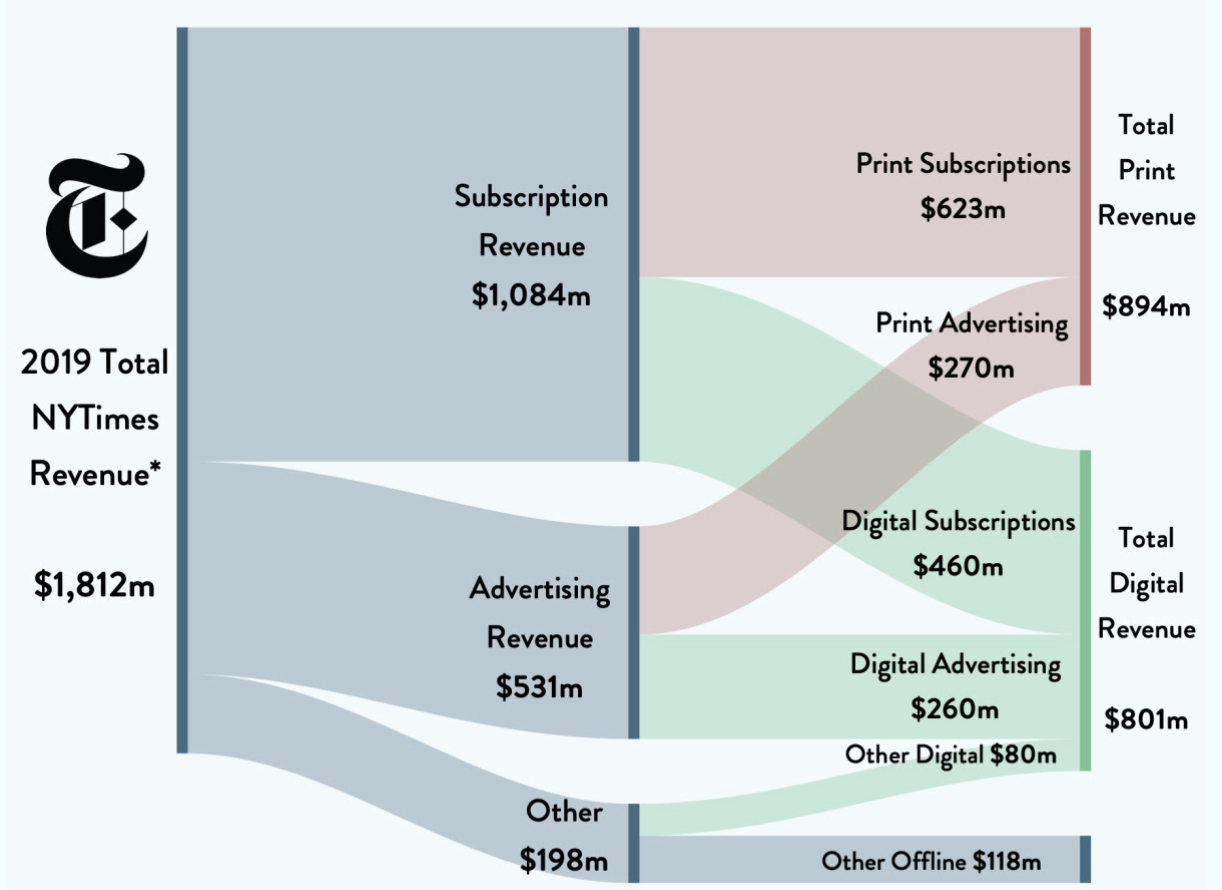}
    \bigskip
    \caption{Revenue breakdown in 2019 (Source: chartr)}
    \label{nyt2}
\end{figure}

\clearpage
\newpage

\section{Targeting Emails}

\begin{figure}[H]
    \centering
    \begin{subfigure}[t]{0.5\textwidth}
        \centering
        \includegraphics[scale=0.4]{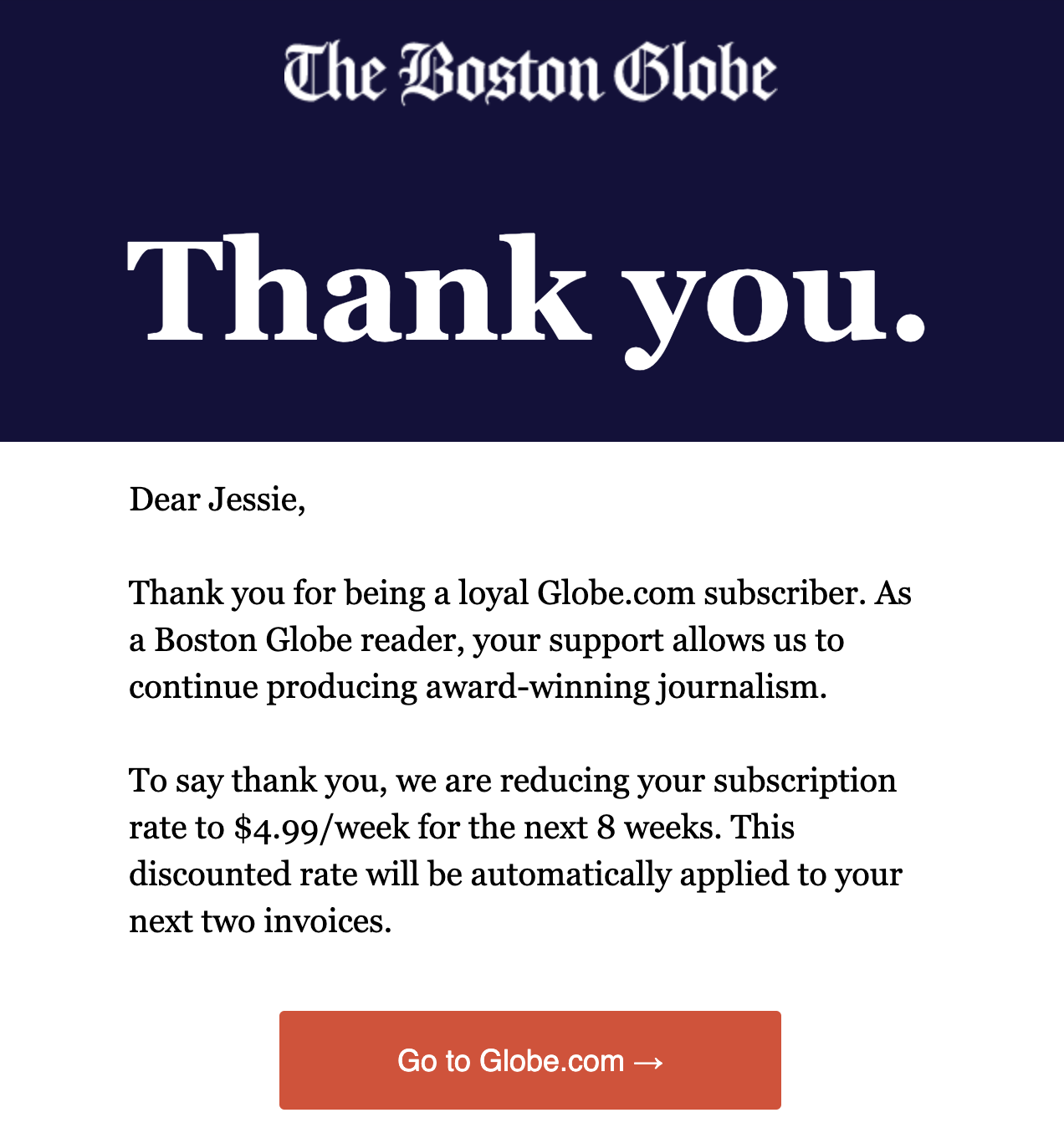}
        \caption{Experiment 1}
        \label{fig1}
    \end{subfigure}%
    ~ 
    \begin{subfigure}[t]{0.5\textwidth}
        \centering
        \includegraphics[scale=0.6]{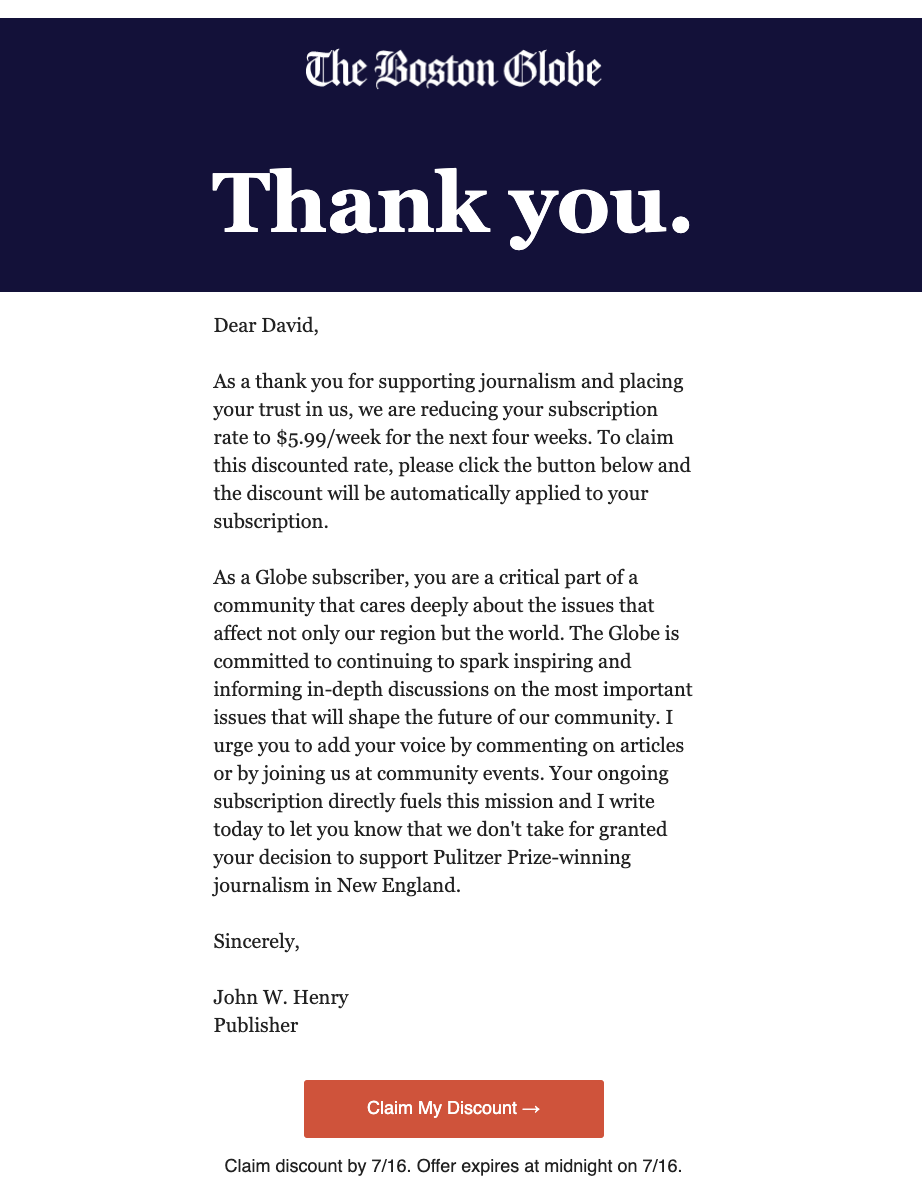}
        \caption{Experiment 2}
        \label{fig2}
    \end{subfigure}
    \caption{Targeting emails. First experiment (left): A sample email sent to targeted subscribers in August 2018, discounts are applied to subscribers automatically. Second experiment (right): A sample email sent to targeted subscribers in July 2019, subscribers have to redeem the offer by clicking on "claim my discount" before the expiration day which is 24 hours after the email was sent. \$5.99/week for 4 weeks is one of the 6 treatment conditions}
    %\label{fig2}
\end{figure}

\clearpage
\newpage

\section{Proof of Propositions}
\label{proof}
\begin{proof}{Proposition 1:}
Consider a case with binary actions. Let $\pi(x) := \pi(1|x)$. We show that the value of a policy as defined on true long-term outcomes $Y$ is identified using the surrogate index $\Tilde{Y}$.
\begin{equation}
\begin{split}
V(\pi) &= \mathbb{E} \left\{ \pi(X_i)Y_i(1) + (1-\pi(X_i))Y_i(0) \right\} \\
&= \mathbb{E} \left\{\pi(X_i) \mathbb{E}[\frac{A_iY_i}{e(X_i)}] + (1-\pi(X_i))\mathbb{E}[\frac{(1-A_i)Y_i}{1-e(X_i)}] \right\}\\
& = \mathbb{E} \left\{ \pi(X_i)\frac{A_iY_i}{e(X_i)} + (1-\pi(X_i))\frac{(1-A_i)Y_i}{1-e(X_i)} \right\}\\
& = \mathbb{E} \left\{ \mathbb{E}[\pi(X_i)\frac{A_iY_i}{e(X_i)} + (1-\pi(X_i))\frac{(1-A_i)Y_i}{1-e(X_i)}|S_i,X_i ]\right\}\\
& = \mathbb{E} \left\{ \pi(X_i)\frac{\mathbb{E}[A_i|S_i,X_i]\mathbb{E}[Y_i|S_i,X_i]}{e(X_i)} + (1-\pi(X_i))\frac{\mathbb{E}[1-A_i|S_i,X_i]\mathbb{E}[Y_i|S_i,X_i]}{1-e(X_i)} \right\}\\
& = \mathbb{E} \left\{ \pi(X_i)\frac{A_i\Tilde{Y_i}}{e(X_i)} + (1-\pi(X_i))\frac{(1-A_i)\Tilde{Y_i}}{1-e(X_i)} \right\}
\end{split}
\end{equation}
$e(X_i)$ is the propensity score. The first line is from the definition of the value of a policy. The second line is because under Assumption \ref{assumption:regular} (ignorability and positivity) we have
\begin{equation}
\begin{split}
&\mathbb{E}[\frac{A_iY_i}{e(X_i)}] = \mathbb{P}(A_i = 1|X_i) \frac{Y_i(1)}{e(X_i)} = Y_i(1)\\
&\mathbb{E}[\frac{(1-A_i)Y_i}{1-e(X_i)}] = \mathbb{P}(A_i = 0|X_i) \frac{Y_i(0)}{1-e(X_i)} = Y_i(0).
\end{split}
\end{equation}
The third line is because $\pi(X_i)$ is a constant. The fourth line is from the law of iterated expectation: We first condition on surrogates and covariates $S_i$ and $X_i$. The fifth line is based on Assumption \ref{assumption:surrogacy} (surrogacy) so the expectation of product can be factorized into the product of expectations. The last line is based on undoing the law of iterated expectations, the definition of surrogate index and Assumption \ref{assumption:comparability} (comparability) as in Equation \ref{si}. The same argument also goes through for multi-action cases.
\end{proof}

\begin{proof}{Proposition 2:}
For policy optimization, consider the case of binary actions, we can see that an optimal policy $\pi^*$ maximizes the average outcome by assigning a subscriber to treatment if and only if the conditional average treatment effect (CATE) for that subscriber is positive (net of the cost of treatment if any):
\begin{equation}
    \begin{split}
\text{argmax}_{\pi} V(\pi) &= \text{argmax}_{\pi} \ \mathbb{E}[ Y(X_i, \pi(X_i))] \\
&=  \text{argmax}_{\pi} \ \mathbb{E}[ \pi(X_i)Y_i(1) + (1-\pi(X_i))Y_i(0) ]\\
&=  \text{argmax}_{\pi} \ \mathbb{E}[ \pi(X_i)(Y_i(1)-Y_i(0)) + Y_i(0) ] \\
&=\text{argmax}_{\pi} \ \mathbb{E}[ \pi(X_i)\tau(X_i) + Y_i(0) ]
\end{split}
\end{equation}

\begin{equation}
\tau(x) := \mathbb{E}[Y_i(1) - Y_i(0)|X_i = x]    
\end{equation}

\begin{equation}
\pi^*(x) = 
\begin{cases} 
      1 & \tau(x) \geq 0 \\
      0 & \tau(x) < 0
\end{cases}.    
\end{equation}

Because the optimal policy depends only on the sign of CATE on the long-term outcome, the policy optimized on surrogate index is valid as long as CATE estimated on the surrogate index is of the same sign as the true CATE.

Following a similar derivation as in the proof of Proposition 1:
\begin{equation}
\label{proof2}
   \begin{split}
        \tau(X_i) &= \mathbb{E} \left[Y_i(1) - Y_i(0)|X_i \right]\\
        &= \mathbb{E}\left[\frac{A_iY_i}{e(X_i)}-\frac{(1-A_i)Y_i}{1-e(X_i)} | X_i \right]\\
        &= \mathbb{E}\left[\mathbb{E}[\frac{A_i Y_i}{e(X_i)}-\frac{(1-A_i)Y_i}{1-e(X_i)}|S_i, X_i]] \right]\\
        &= \mathbb{E}\left[\frac{\mathbb{E}[A_i|S_i,X_i]\mathbb{E}[Y_i|S_i,X_i]}{e(X_i)} - \frac{\mathbb{E}[1-A_i|S_i,X_i]\mathbb{E}[Y_i|S_i,X_i]}{1-e(X_i)}|X_i]\right]\\
        &= \mathbb{E}\left[\frac{A_i\Tilde{Y}_i}{e(X_i)}-\frac{(1-A_i)\Tilde{Y}_i}{1-e(X_i)}|X_i \right]\\
   \end{split}
\end{equation}
The surrogate index can be used to construct an unbiased estimator of CATE, therefore, it can be used for policy learning. 
\end{proof}

%\subsection{Loss Bound when Surrogacy Assumption is Violated} \label{loss}

\begin{proof}{Proposition 3:}
There is no loss in policy value if surrogate-index-based policy identifies the true optimal action, i.e., $a^*(X) = \tilde{a}^*(X)$. Note that this can be true even when the CATE is biased. When the surrogate-index-based policy doesn't identify the true optimal actions, the loss in policy value is the difference in outcome under the true optimal action and the one identified by the policy, i.e., ${\tau}_{a^*\tilde{a}^*}(X)$. Therefore, the total loss in policy value when integrating over the distribution of $X$ is:

\begin{equation}
    \int_X {\tau}_{a^*\tilde{a}^*}(X) \cdot \mathbbm{1}_{\{a^*(X) \not = \tilde{a}^*(X)\}} \ dF(X) 
    %\{ \sgn(\tCATE(X)) \not = \sgn(\CATE(X))\}
\end{equation}

When Assumption 2 (surrogacy) is violated, the CATE estimated using surrogate index is biased. In binary cases, \cite{athey2019surrogate} showed that the bias on ATE is bounded by $\bar{b}$:
\begin{equation}
    |b| \leq \left(\frac{\text{var}(Y_i)}{\text{var}(A_i)}\cdot (1-R^2_{Y|S}) \cdot (1-R^2_{A|S}) \right)^{\frac{1}{2}}:= \bar{b}
\end{equation}
where $R^2_{Y|S}$ and $R^2_{A|S}$ is the $R^2$ of the regression of long-term outcome on surrogates (in the historical dataset), and the regression of actions on surrogates (in the experimental dataset), respectively. Similarly, the bias on CATE is bounded by $\bar{b}_X$:
\begin{equation}
    |b_X| \leq \left(\frac{\text{var}(Y_i|X_i)}{\text{var}(A_i|X_i)}\cdot (1-R^2_{Y|S,X}) \cdot (1-R^2_{A|S,X}) \right)^{\frac{1}{2}}:= \bar{b}_X
\end{equation}
because an optimal policy assigns actions based on the sign of true CATE, as long as the
CATE estimated on surrogate index has the same sign as the true CATE, there’s no loss on
the value of policy. When the signs are different, the loss is the true CATE, it follows that the total loss in policy value is bounded above by:
\begin{equation}
    \int_X (\bar{b}_X - |\tilde{\tau}(X)|) \cdot \mathbbm{1}_{\{\bar{b}_X - |\hat{\tau}(X)| > 0\}} \ dF(X) 
\end{equation}
where $\tilde{\tau}(X)$ is the CATE estimated with the surrogate index. 
\end{proof}

\clearpage
\newpage

\section{Supplementary Analyses}

\subsection{Churn Prediction} \label{churn prediction}

The dataset we have includes demographic, transaction history and browsing history for all digital only subscribers from 2010/12/16 to now. We first pick a date and use all the information before that date to predict outcomes (whether a given subscriber churned or not) that happened within six months after that date. We picked 2018/01/30 because it gives us the most recent information before targeting subscribers in the first experiment, although model performance is robust to other dates that we picked. 

We select active subscribers defined by the company, it includes all subscribers who are currently active, in grace period, or in temporary stop.\footnote{Most common reason for this is traveling.} Then we construct features from transaction history using frequency and recency by each transaction type, which are standard features in the churn prediction literature \citep{lemmens2006bagging}. Frequency is the number of times a certain transaction type occurred, and recency is the first and last time a certain transaction type occurred compared with the date we picked (in days). Then we count the number of articles read in the last week, month, 3 months and 6 month to measure the level of engagement. We also extracted how many articles a subscriber read in each section on the newspaper's website over time, although this content consumption information is not used for churn prediction, we use it for policy learning. We look at churn that happened between 2018/01/30 and 2018/07/18 to get the outcome labels, if a churn happened it's coded as 1, and 0 otherwise. We handle missing data in the following way: if a feature is a measure of recency, then missing means that a certain type of event has not happened yet, so we impute a large positive number 1000 (a positive number means it is in the future) and create a separate column indicating if that value is missing (1 or missing, 0 for not missing). If a feature is categorical, we create ``missing'' as a new category. Altogether we have 183 features. We also removed recent subscribers whose tenure is less than 60 days and who hasn't opened any emails sent by the company in the last 6 months. The reason is that recent subscribers are likely to be on an introductory rate which is already discounted, we don't want to give them more discounts. 

Then we build a classification model to predict the churn risk for each subscriber by combining information from three different sources: demographics (e.g., zip code), transaction history (credit card status, credit card expire date and transaction type, including auto notice, auto renew, refund, billing change, complaint, expire, end of grace period, payment cancel, payment declined, start, stop, etc., and associated time stamp, and a source and reason code associated with each transaction), and browsing history on the newspaper's official website (number of articles read and associated time stamp, article section, article headline) from 2010/12/16 to 2018/07/18. We trained and compared a wide range of classification algorithms. Among the models we trained, gradient boosted decision trees (XGBoost) \citep{chen2016xgboost, chen2015xgboost} have the best out-of-sample performance measured by AUC (area under the curve). See Table \ref{tab15} for a comparison. We have an overall out-of-sample accuracy of 97\%\footnote{This is not surprising given the class labels are highly imbalanced. There are about 4\% churn rate in the data, the overall accuracy is most from correctly predicting non-churners.}, and precision of 94 \%\footnote{Precision is the proportion of actual churners among predicted churners. It means that when we predict a subscriber to be a churner, 94\% of the time we are correct.}. However, the recall is low at 23\%\footnote{Recall is the proportion of predicted churners among actual churners. It means that among all the actual churners we correctly identified 23\% of them.} suggesting that we might have missed some important signals when constructing features or the information is simply unobserved.

As in many classification problems, we need to trade-off the cost of false positive and false negative. In our setting, a false positive is a non-churner classified as a churner, and a false negative is a churner classified as a non-churner. The cost of a false positive is the cost of the discount. Since the subscriber is not going to churn, the firm wasted $(\$6.93-\$4.99) \times 8 = \$15.52$ per targeted subscriber. The cost of a false negative is harder to evaluate because it depends on how soon the churn happened and how long she would have stayed if she had been targeted with the discount. Assuming a churner churned in 2 month, the revenue collected without the discount is $\$6.93 \times 8 = \$55.4$, if the churner would stay for an extra month if she received the discount, then the revenue collected would be $ \$4.99 \times 8 + \$6.93 \times 4= \$67.6$, therefore the cost would be $\$12.2$.

Table \ref{tab15} shows the prediction performance of a menu of classification models, and we can see that XGBoost outperforms other models by a significant margin. Table \ref{tab16} is the confusion matrix of the performance of XGBoost on a test sample of size 8000. We can see that the precision is very high (we get 60 out of 64 right when we predict someone to be a churner), but the recall is low (we correctly predict 60 out of 265 real churners). Table \ref{tab17} is the top 20 features that are predictive of churn, we can see that credit card information is very important, the company also mentioned that a big number of subscribers (over 25\%) churn is from an expired credit card (they send notification emails to tell the subscribers if their cards are about to expire). Auto renew and billing change information are also important, so is the level of engagement as measured by the number of articles read last week, month, and 6 months.

\begin{table}[!htbp] \centering
  \caption{Predictive model performance}
  \label{tab15}
  \bigskip
\begin{tabular}{ |c|c| } 
 \hline
 \hline
 Model & AUC \\
 \hline
 Logistic Regression & 0.7557 \\
 Support Vector Machine & 0.5824 \\
 Random Forest & 0.5669 \\ 
 XGBoost & 0.9384\\
 \hline
\end{tabular}
\end{table}

\bigskip

\begin{table}[!htbp] \centering
 \caption{Confusion matrix for XGBoost on test data}
  \label{tab16}
  \bigskip
 \begin{tabular}{|c|c|c|} 
 \hline
 predicted/actual & 0 & 1\\
 \hline
0 & 7731 & 205 \\
 \hline
1 & 4 & 60 \\
 \hline
\end{tabular}
\end{table}

\begin{table}[!htbp] \centering
 \caption{Relative feature importance in XGBoost (top 20)}
  \label{tab17}
  \bigskip
\begin{tabular}{|c|c|} 
\hline
feature & relative importance\\
\hline
credit\_card\_statusa       &    100.000\\
credit\_card\_statusi       &    66.728\\
last\_autorenew                &   39.728\\
cc\_expire\_dt                  &  31.951\\
last\_billingchg\_reasonremovecc  & 23.667\\
first\_billingchg\_reasonremovecc  &18.981\\
last\_start\_tenure               &7.919\\
credit\_card\_typeu                &6.016\\
original\_tenure                 &5.786\\
last\_billingchg                   &5.252\\
first\_autorenew                   &4.331\\
last\_expire                       &3.954\\
first\_billingChg                  &3.588\\
last\_6month                       &3.501\\
last\_week                         &3.346\\
last\_month                        &3.281\\
num\_autorenew                     &2.621\\
num\_billingChg                    &2.252\\
num\_pymtdecline                   &1.731\\
first\_pymtdecline                 &1.648\\
\hline
\end{tabular}
\end{table}

\clearpage
\newpage
\subsection{Design Policy} \label{design policy}
We obtain the design or behavior policy, which is the targeting policy we implement in the first experiment, by garbling the predicted risk score from the XGBoost with random noise generated from a normal distribution.\footnote{It is the best performing model for churn prediction, see Appendix \ref{churn prediction} for more details.} The key idea is that we are treating subscribers with higher risk of churn with higher probability, but allow everyone to be either treated or not with positive probability.

The reason that we base the design policy on predicted risk score is twofold. First, because churn risk is an outcome bounded between 0 and 1, it provides an upper bound on how big the beneficial treatment effect\footnote{Beneficial means when treatment effect is in the direction that moves the outcome in a desirable direction.} can be without any further assumptions. For instance, if a subscriber has a predicted risk of 0.1, it means that the discount will \textit{at most} lower her risk by 0.1, on the other hand, if a subscriber has a predicted risk of 0.9, the discount can lower her risk by \textit{up to} 0.9, provided that the model is well calibrated. So it's reasonable to treat subscribers with higher risk with higher probability without any additional information. This approach can also be interpreted as treating subscribers based on an upper confidence bound (UCB) of the beneficial treatment effect with minimal assumptions. Second, if risk of churn is indeed positively correlated with treatment effect, this approach lowers regret compared with a uniform policy which is the most typical exploration policy, it also gives us more precision to learn the policy at a region that matters the most (the region where the treatment effects are the highest) because we are assigning more subscribers in this region to treatment. We conduct a simple simulation analysis to further illustrate this. The result shows that, under bounded outcome while both policies recover the true ATE well, the design policy that assigns subscribers to treatment with probability proportional to her churn risk has lower regret compared with a uniformly at random policy, and this is true under very general conditions.  

The reason we added noise to predicted risk is also twofold. First, we want to explore more around the predicted risk score. Without the noise, the targeting policy would reduce to a version that is the common practice, which is to target based on predicted outcome level, which is the churn risk in this context \citep{blattberg2008database}. Some exploration allows us to learn the treatment effect at regions that our prior thinks the effect is low, that is, subscribers with medium to low predicted risk of churn. This allows us to learn an optimal policy even when our prior is wrong. Second, to use the inverse propensity score for off-policy evaluation and learning, we need all subscribers to have a positive probability of being in all conditions\footnote{Note that this condition is usually not satisfied using the common practice. Suppose the targeting policy is to treat subscribers who are in the top 5\% of churn risk, then 95\% of subscribers have zero probability of receiving the treatment by design. }. Even when this condition is satisfied in principle, the variance of the counterfactual policy evaluation is very large and unstable when some of the probabilities are very close to zero \citep{dudik2014doubly}. After adding noise, we essentially make the propensity scores more smooth, that is, the probability of receiving the treatment for the top churners gets lower, and the probability for the bottom churners gets higher. It ensures that everyone has a propensity score that is bounded away from 0 and 1.

\begin{figure}[!htb]
    \centering
    \includegraphics[scale=0.4]{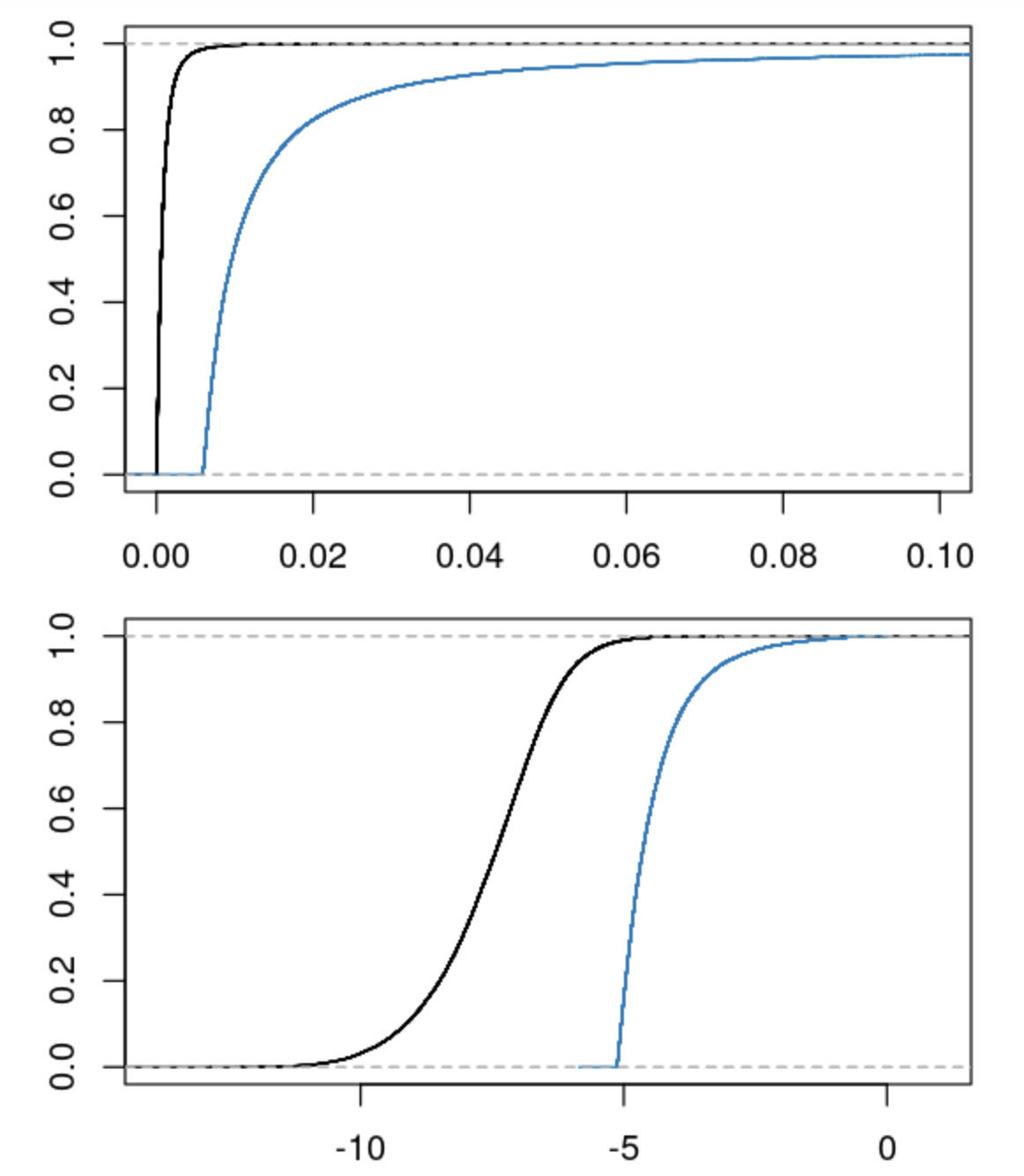}
    \caption{CDF of risk scores (black) and treatment probability under the design policy (blue) on regular (top) and log scale (down): most subscribers have risk score close to zero, the design policy increases the treatment probability for those subscribers, but a big majority (over 80\%) still has a treatment probability below 2\%, the treatment probability is also capped at 50\% to ensure sufficient exploration.}
    \label{fig3}
\end{figure}

More formally, let $R_i \in (0,1)$ be the predicted risk for subscriber $i$. A stochastic targeting policy $\pi$ is defined as:
\[\pi: X \to (0,1)\].
it's a mapping from $X$, which is the covariate space of subscribers, to an open probability interval. Note that it's important for a design policy to be stochastic, meaning that every subscriber under the design policy has to have nonzero probability of both receiving the treatment \textit{and} the control, so the interval should be open on both ends. The policy we want to evaluate can be both stochastic and deterministic, we only require the design policy to be stochastic. Because a policy is just the probabilities that subscribers receive the treatment, we can think of it as a vector of propensity score. Given the predicted risk score $S$, our design policy $\pi_D$ is given by:
\begin{align*}
\pi(1|R_i) &= \Pr(R_i - \epsilon \geq \tau)\\
& =\Pr(\epsilon \leq s_i - \tau)\\
& = F_\sigma(R_i-\tau)
\label{eq6} \end{align*}
where $\epsilon \sim N(0,\sigma^2)$ is the random noise added, $F$ is the CDF of $\epsilon$. $\sigma$ controls the amount of exploration and $\tau$ is a constant threshold that controls the total number of subscribers treated. This policy is fully characterized by the choice of $\sigma$ and $\tau$. In the first experiment the firm wants to send discounts to about 1,000 subscribers. In the design policy we implemented we have $\sigma=0.003$, $\tau=0.0068$. And we cap the probability of receiving the treatment at 50\% for all subscribers. See Figure \ref{fig3} for the CDF of treatment probability before and after adding the noise (it's just the raw predicted churn risk before adding noise). 

To make the idea more concrete, we conducted a simple simulation to compare the performance of a uniformly at random policy and a design policy that assigns subscribers with higher churn risk to treatment with higher probabilities. We are particularly interested in (1) how good the outcomes are in the experiment and (2) how well the learning is.

Consider the following data generating process: let $0 < Y_i(0) < 1$ be the baseline churn risk for subscriber $i$ without any interventions, this is essentially the output of our churn classification algorithm described in the previous section, and because of this we treat it as observable for all subscribers. Now suppose the intervention lowers churn risk, without any further assumptions we can draw $Y_i(1)$ uniformly from the interval $(0, Y_i(0))$, that is, we know the post treatment outcome $Y_i(1)$ has to be bounded above by $Y_i(0)$ and below by 0. Now we have the full schedule of potential outcomes, we can simulate two types of experiments: assigning subscribers to treatment with probability 0.01 (we call this the uniform policy), and assigning subscribers to treatment with probability proportional to churn risk (we call this the design policy) but keep the total fraction of treated subscriber fixed at 0.01. We compare (1) what's the average churn under uniform and design policy and (2) what's the estimated treatment effects under uniform and design policy (because we have the full schedule of potential outcome we can compare it with the ground truth). 

\begin{figure}
    \centering
    \includegraphics[scale=0.6]{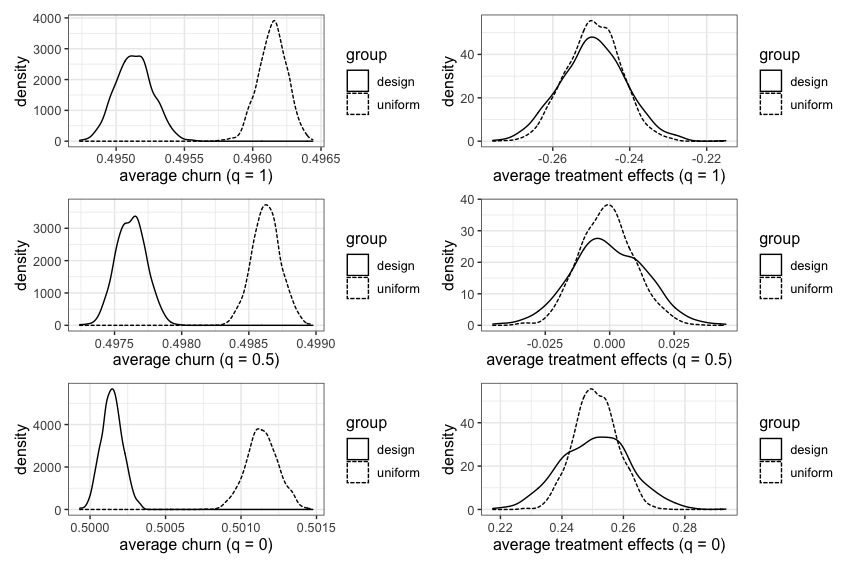}
    \bigskip
    \caption{Design vs. uniform policy on churn rate and ATE}
    \label{fig27}
\end{figure}

Similarly, if the intervention increases churn risk, we draw $Y_i(1)$ uniformly from the interval $(Y_i(0),1)$, that is, the post treatment outcome $Y_i(1)$ is bounded below by $Y_i(0)$ and above by 1. More generally, we can let the treatment effect for a given subscriber be negative with probability $q$ and positive with probability $1-q$ and repeat the procedure, $q$ captures the fraction of subscribers on whom the intervention has a negative treatment effect (we think in practice $q$ should be quite large). We do 1000 repetitions according to policy with $q = 0, 0.5, 1$ and report the results in Figure \ref{fig27}. We can see that the design policy has lower churn rate in all cases, and both design policy and uniform policy recover the true average treatment effect (ATE). To further extend this analysis, we can allow the distributions of treatment effects to take different shapes similar to the simulation studies conducted in \cite{misra2019dynamic}. 

\clearpage
\newpage

% \subsection{Survival Curves}
% \label{survival curves}

\subsection{Treatment Effects}
\label{treatment effects}

\subsubsection{Experiment 1}
\begin{figure}[h]
    \centering
        \includegraphics[scale=.65]{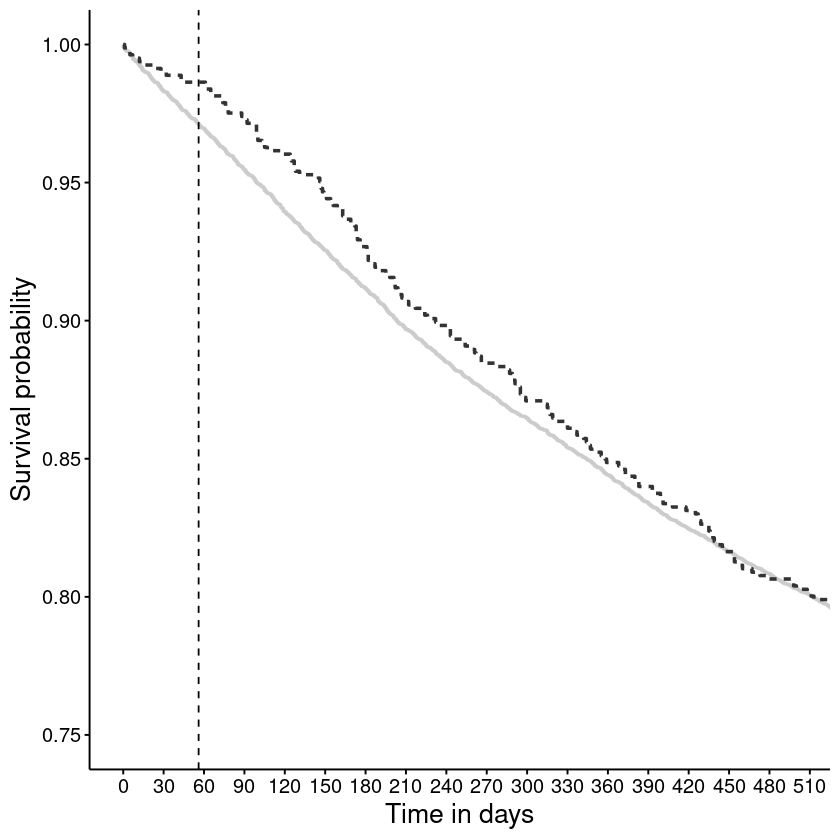}
    \caption{Empirical survival curve from the first experiment}
    \label{fig4}
\end{figure}

We plot the empirical survival curves of subscribers in the first experiment in Figure \ref{fig4} using data from August 2018 -- February 2020 (the dashed line is the treatment group). The first thing to notice is that the survival rate is relatively high, about 80\% of subscribers at the beginning of the experiment remain subscribers 18 months later. Second, there is a gap between treatment and control group. Note that the treatment and control groups are not directly comparable because the treatment group has subscribers with higher churn risk, so we would expect the dashed line to be below the solid line without the treatment, the fact that the dashed line is mostly above the solid line shows the treatment has a big effect in reducing churn.

\begin{figure}[h]
    \centering
    \includegraphics[scale=.7]{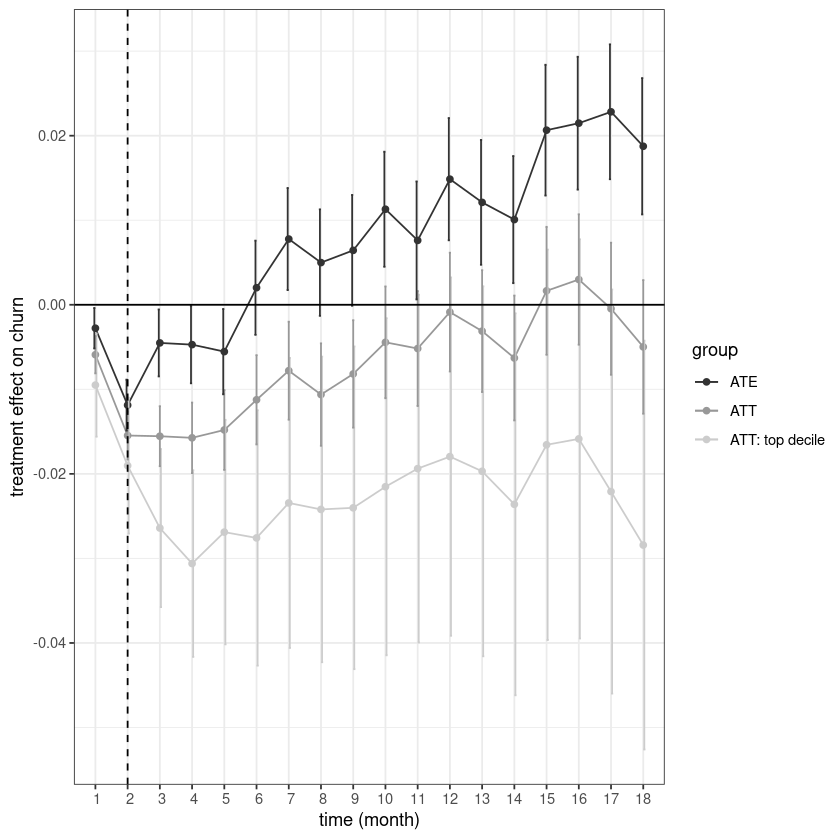}
    \caption{Treatment effects on churn over time in the first experiment. ATE is the average treatment effect on all subscribers, ATT is the treatment effect on treated subscribers, and ATT top decile is the ATT on subscribers with risk of churn in the top decile. The discount ends in month two (dashed vertical line).}
    \label{fig5}
\end{figure}

\begin{figure}[h]
    \centering
    \includegraphics[scale=.7]{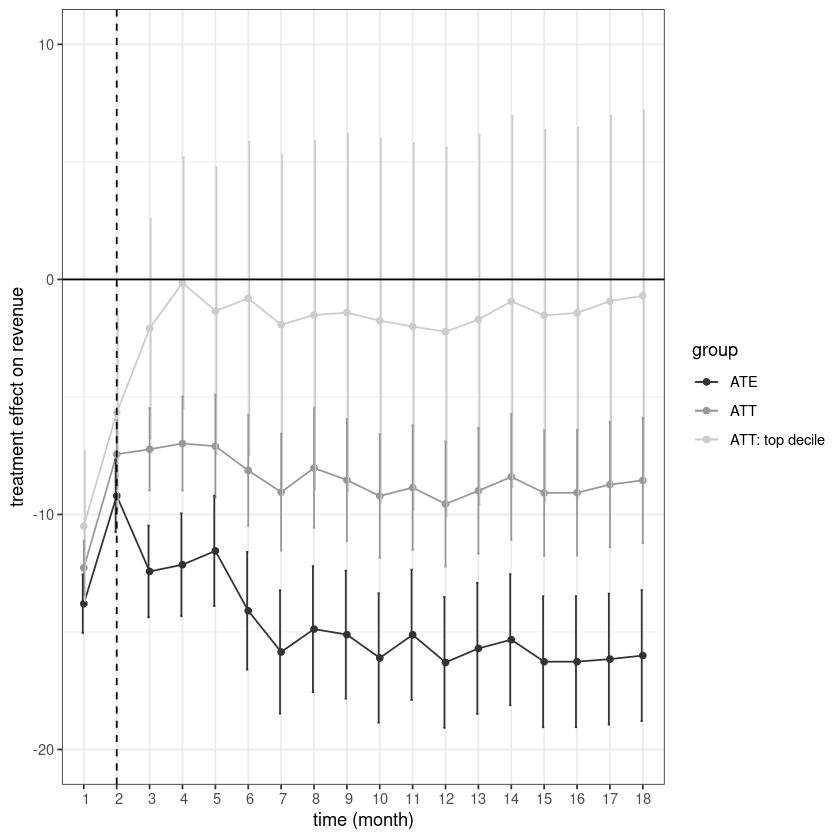}
    \caption{Treatment effect on revenue over time in the first experiment. ATE is the average treatment effect on all subscribers, ATT is the treatment effect on treated subscribers, and ATT top decile is the ATT on subscribers with risk of churn in the top decile. The discount ends in month two (dashed vertical line).}
    \label{fig6}
\end{figure}

We plot the average treatment effect (ATE) and the average treatment effect on the treated (ATT) over time using churn and revenue as outcomes in Figures \ref{fig5} and \ref{fig6}.
Note that we do not necessarily expect beneficial effects here (i.e., policy optimization can succeed even if the ATE and ATT are not beneficial). This is particularly true for the ATE, which is averaging over all subscribers, many of whom have low risk of churn.
When using churn as outcome, ATE has the smallest effect size and is marginally significant at month 3 (the discount ends after month 2).
The ATT stays negative but only marginally significant after month 10. That the ATT is bigger than the ATE in effect size suggests that our design policy assigns more subscribers for whom the discount tends to have a beneficial effect to treatment, which is better than a uniformly random policy. The ATT on the subset of subscribers with the highest risk shows the biggest effect which provides supportive evidence to our prior and the choice of design policy. When looking at revenue, we see that the treatment effects are mostly negative 18 months after the experiment. This is likely due to two factors: (1) we might need to wait longer for a positive effect, which is consistent with our focus on long-term outcomes, (2) our design policy is not targeting the optimal set of subscribers, if we did so, as we will show in the policy learning section, the 18-month revenue impact will be positive.

\subsubsection{Experiment 2}
\begin{figure}[h]
    \centering
    \includegraphics[scale=0.75]{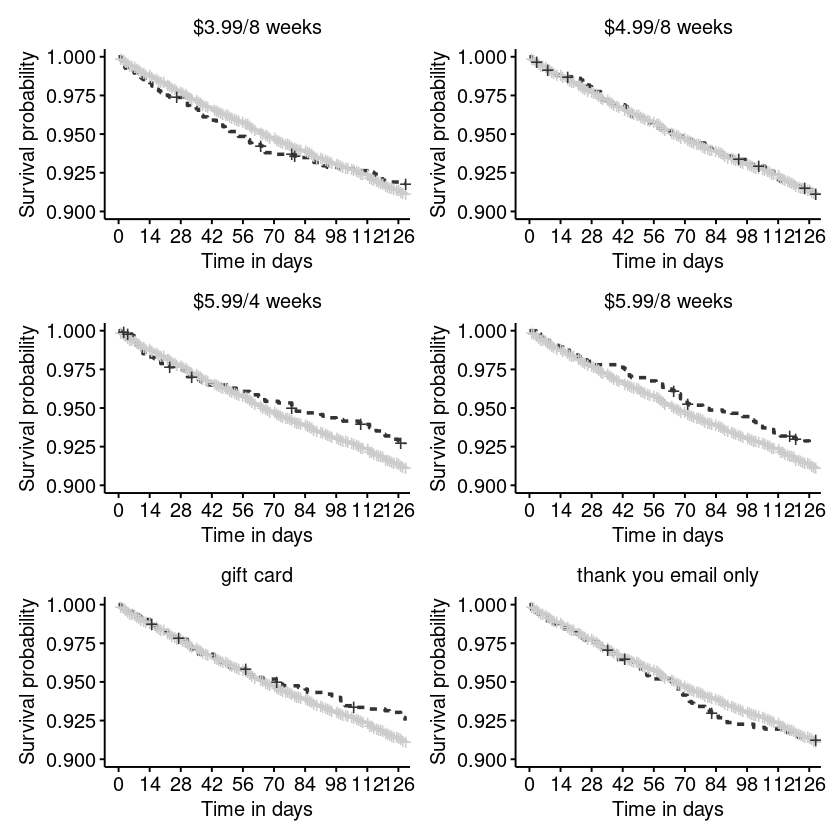}
    \caption{Empirical survival curve in the second experiment by treatment conditions. Dashed (grey) curve is the treatment (control) group. %Vertical dashed line indicates when the discount ends (there's no ending date for gift card and thank you email only condition)
    } 
    \label{fig:cohort2_survival}
\end{figure}

We plot the survival curves of subscribers in the second experiment in Figure \ref{fig:cohort2_survival} using data from July 2019 to February 2020 (dashed lines are treatment groups). Surprisingly, \$5.99/4 weeks and \$5.99/8 weeks, which give the smallest discounts, have the biggest treatment effect on churn reduction. This, in turn, translates into the bigger effects on revenue. 

\begin{figure}[h]
    \centering
    \includegraphics[scale=0.8]{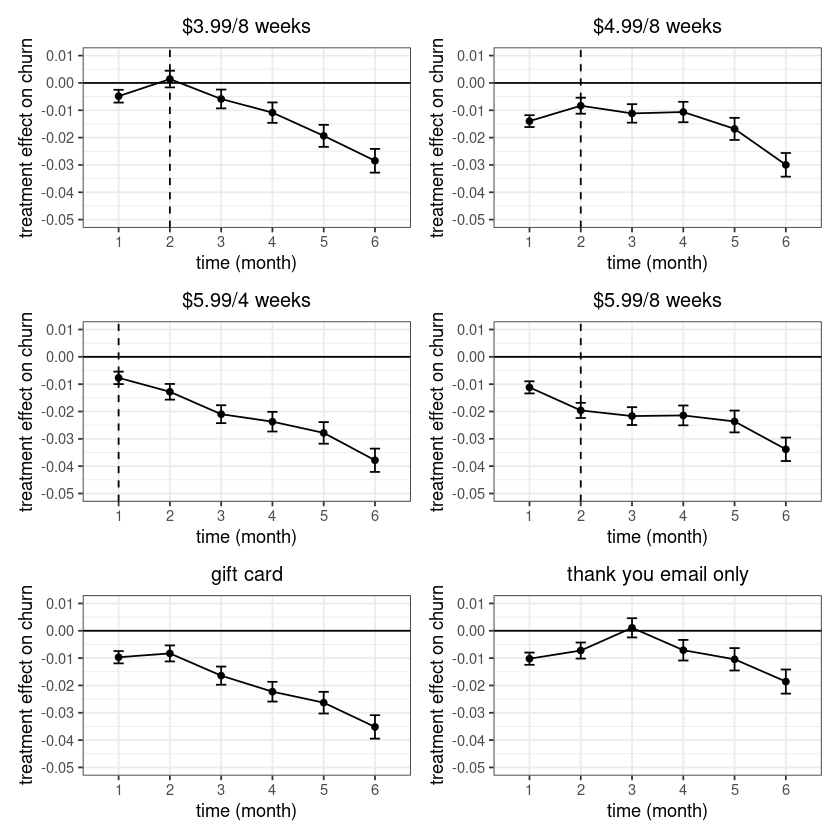}
  \caption{ATT on churn in the second experiment by treatment conditions. Vertical dashed lines indicate when the discount expires. Vertical dashed lines indicate when the discount ends (when applicable).}
    \label{fig14}
\end{figure}

\begin{figure}[h]
    \centering
    \includegraphics[scale=0.8]{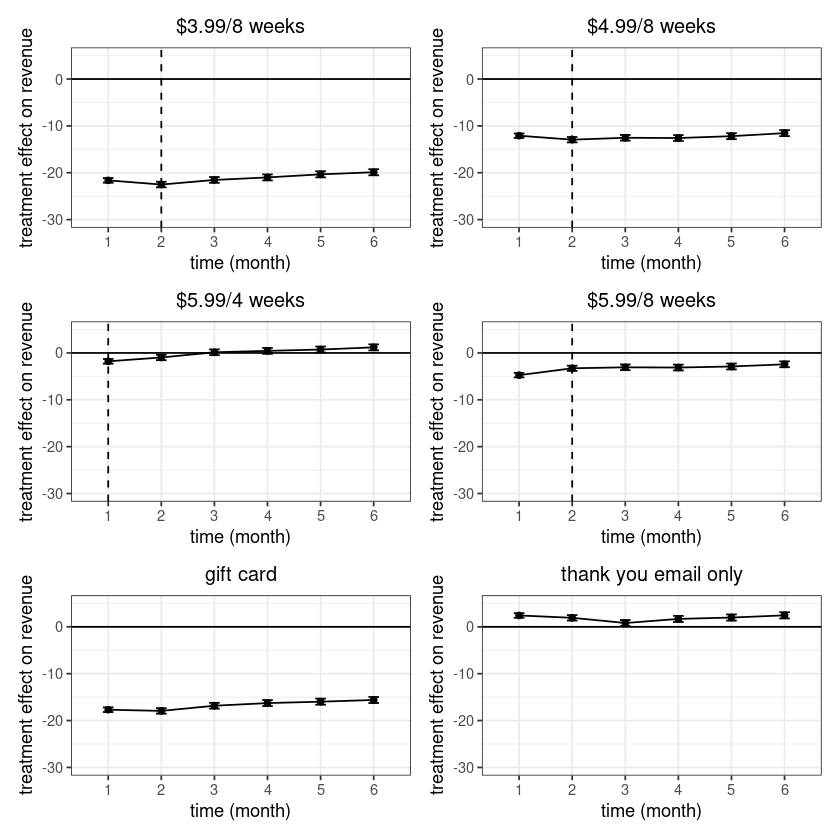}
  \caption{ATT on revenue in the second experiment by treatment conditions. Vertical dashed lines indicate when the discount ends (when applicable).}
    \label{fig15}
\end{figure}

The ATT for churn and revenue are reported in Figures \ref{fig14} and \ref{fig15} by treatment conditions. \$5.99/4 weeks and \$5.99/8 weeks, which give the smallest discounts, have the biggest treatment effect on churn reduction. This also shows up on the revenue plot. We can see that it takes much shorter for \$5.99/4 weeks to break even compared with other conditions (except for email only condition which doesn't have cost).

We also provide some validation of estimated treatment effects by regressing churn and revenue on the interaction between treatment and treatment probability estimated. There is a significantly higher effect on subscribers that are predicted to have a bigger effect using data from the first experiment (Table \ref{tab:interaction_with_score}).\footnote{We reported ATT in the table using inverse probability weights in the regression.} 

\begin{table} \centering 
  \caption{Interaction between treatment and treatment probability} 
  \label{tab:interaction_with_score}
 
\begin{tabular}{@{\extracolsep{5pt}}lcc} 
\\[-1.8ex]\hline 
\hline \\[-1.8ex] 
 & \multicolumn{2}{c}{\textit{Dependent variable:}} \\ 
\cline{2-3} 
\\[-1.8ex] & churn & revenue \\ 
\hline \\[-1.8ex] 
 3.99/8 weeks & $-$0.016$^{***}$ (0.003) & $-$22.032$^{***}$ (0.279)\\ 
 % & (0.003) \\ 
  & \\ 
 4.99/8 weeks & $-$0.005 (0.003) & $-$14.055$^{***}$ (0.280) \\ 
%  & (0.003) \\ 
  & \\ 
 5.99/4 weeks & $-$0.022$^{***}$ (0.003) & $-$1.996$^{***}$ (0.279)\\ 
 % & (0.003) \\ 
  & \\ 
 5.99/8 weeks & $-$0.025$^{***}$ (0.003) & $-$4.994$^{***}$ (0.279)\\ 
 % & (0.003) \\ 
 & \\ 
 gift card & $-$0.020$^{***}$ (0.003) & $-$18.214$^{***}$ (0.280) \\ 
%  & (0.003) \\ 
 & \\ 
 thank you email only & $-$0.012$^{***}$ (0.003) & 0.905$^{***}$ (0.278)\\ 
 % & (0.003) \\ 
  & \\ 
 treatment prob & $-$0.005$^{***}$ (0.001) & 0.280$^{***}$ (0.102)\\ 
%  & (0.001) \\ 
  & \\ 
 3.99/8 weeks $\times$  treatment prob & $-$0.0002 (0.002) & $-$0.083 (0.144) \\ 
%  & (0.002) \\ 
  & \\ 
 4.99/8 weeks $\times$  treatment prob & $-$0.003$^{*}$ (0.002) & 0.116 (0.146)\\ 
 % & (0.002) \\ 
  & \\ 
 5.99/4 weeks $\times$  treatment prob & $-$0.003 (0.002) & 0.530$^{***}$ (0.145)\\ 
%  & (0.002) \\ 
  & \\ 
 5.99/8 weeks $\times$  treatment prob & $-$0.006$^{***}$ (0.002) & 0.504$^{***}$ (0.145) \\ 
%  & (0.002) \\ 
  & \\ 
 gift card $\times$  treatment prob & $-$0.006$^{***}$ (0.002) &  0.152 (0.146)  \\
 % & (0.002) \\ 
  & \\ 
 thank you email only $\times$  treatment prob & 0.003$^{*}$ (0.002) & $-$0.332$^{**}$ (0.145) \\ 
%  & (0.002) \\ 
  & \\ 
 constant & 0.105$^{***}$ (0.002) & 120.849$^{***}$ (0.197)\\ 
%  & (0.002) \\ 
  & \\ 
\hline \\[-1.8ex] 
Observations & 95,554 & 95,554\\
\hline \\[-1.8ex] 
\textit{Note:}  & \multicolumn{2}{r}{$^{*}$p$<$0.1; $^{**}$p$<$0.05; $^{***}$p$<$0.01} \\ 
\end{tabular} 
\end{table}

\clearpage
\newpage
\subsection{Policy Interpretation}
\label{interpretation}

\begin{figure}[!htb]
    \centering
    \includegraphics[scale=0.27]{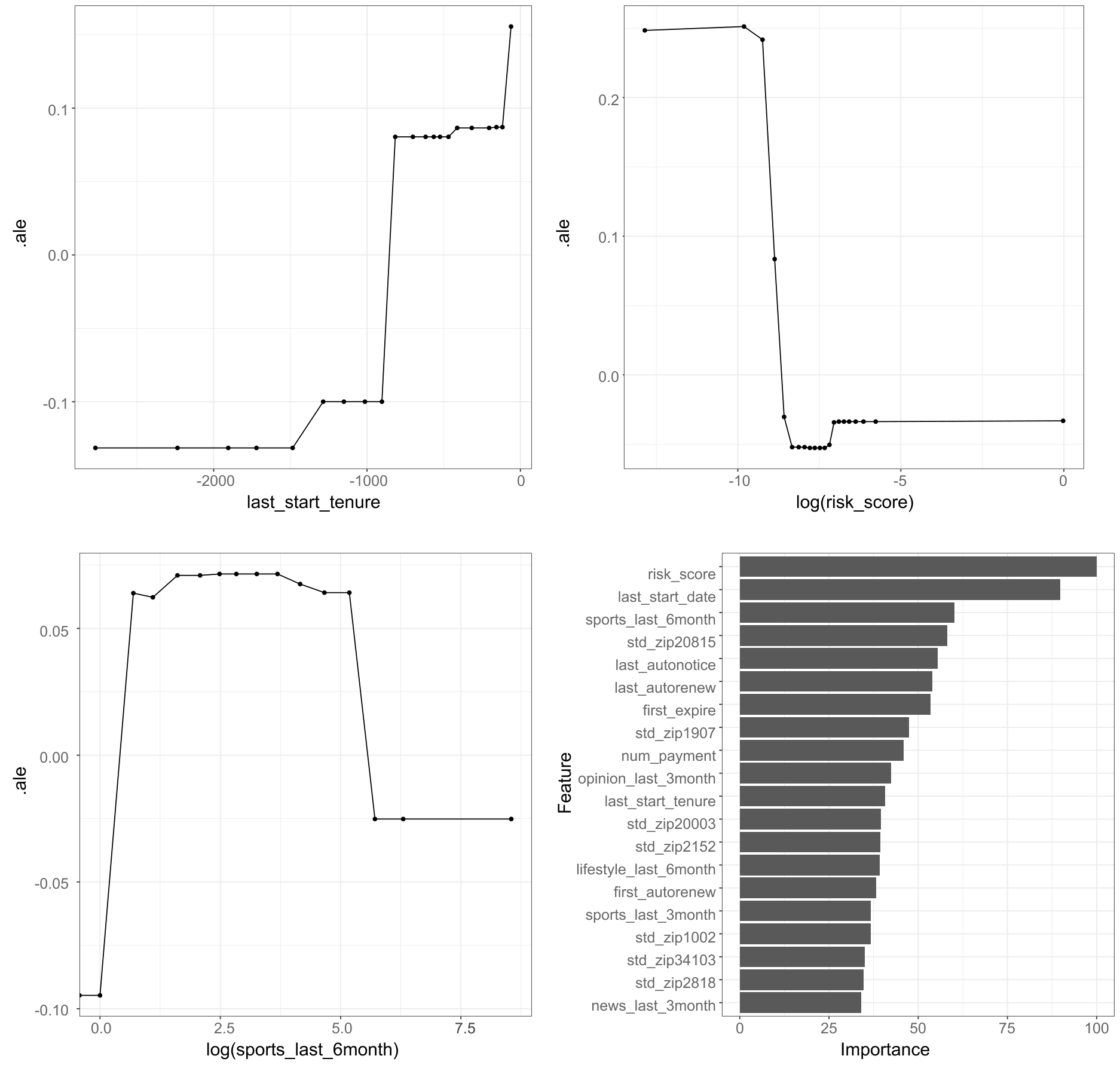}
    \caption{ALE and feature importance} 
    \label{fig11}
\end{figure}

To better understand the learned policy, we use some measures of which covariates are most important. First, we examine (Figure \ref{fig11}, lower right) a standard feature importance measure based on permutation \citep{chen2015xgboost}. This feature importance measure works by calculating the increase of the model prediction error after randomly permuting the feature.\footnote{A feature is more important if permuting its values increases the model error, because the model relied more on the feature for the prediction. A feature is less important if permuting its values keeps the model error unchanged, because the model ignored the feature for the prediction.}
The top 3 features are risk score (the pre-treatment risk of churn), tenure (how long a subscriber has been a subscriber) and number of sports articles read in the last 6 month (a measure of content consumption and how active a subscriber is on the website). Zip code and other content and account info also show up in the top 20 features.

Second, we examine accumulated local effects (ALE)\footnote{ALE is similar to partial dependence but takes feature correlations into account: instead of averaging over distribution of other features in the whole dataset, ALE averages over the distribution of other features conditional on the value of a focal feature \citep{apley2016visualizing}.} for the top three features (Figure \ref{fig11}). ALE shows how treatment probability changes when we vary the values of risk score, tenure and number of sports articles read, respectively. The optimal policy treats subscribers with shorter tenure (more recently registered subscribers) with higher probabilities. The relationship between treatment probability and number of sports articles read is not monotone: the probability is low for very inactive and active subscribers but higher for subscribers in between. The relationship with risk score is interestingly also not monotone, for subscribers with the highest risk scores the treatment probabilities are higher, this is consistent with our prior. But for some subscribers with very low risk scores, the treatment probabilities are even higher. This also highlights the risk of targeting solely based on risk scores. 

\clearpage
\newpage

\subsection{Non-stationarity}
\label{non-stationarity}

\begin{figure}[!htb]
\centering
  \begin{subfigure}[t]{.6\textwidth}
    \centering
    \includegraphics[width=\linewidth]{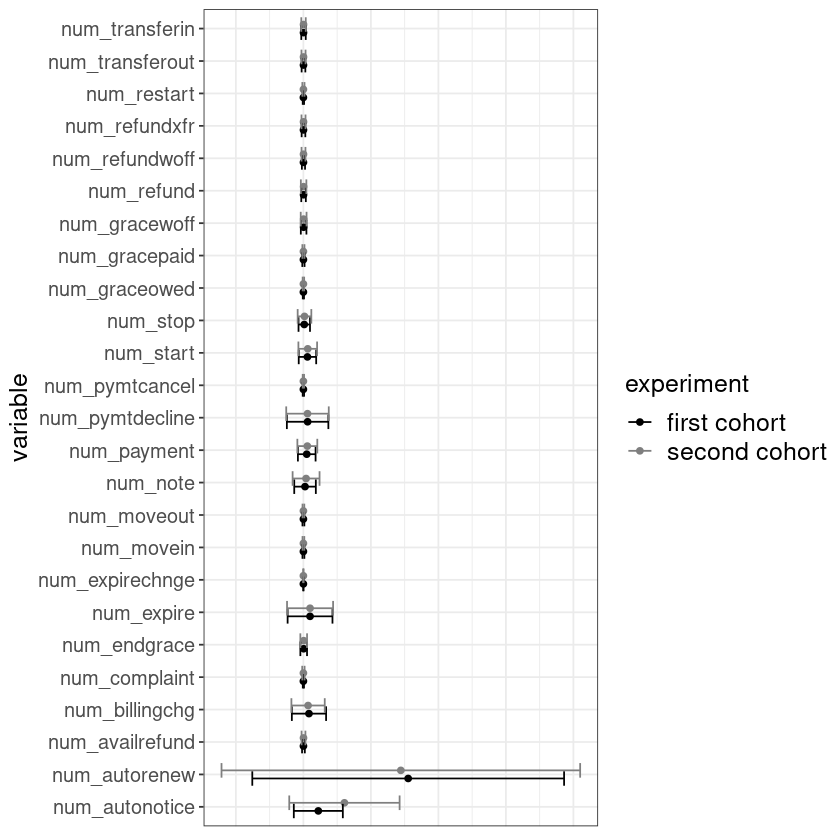}
    \caption{Account Activity}
  \end{subfigure}%
  \hfill
  \begin{subfigure}[t]{.6\textwidth}
    \centering
    \includegraphics[width=\linewidth]{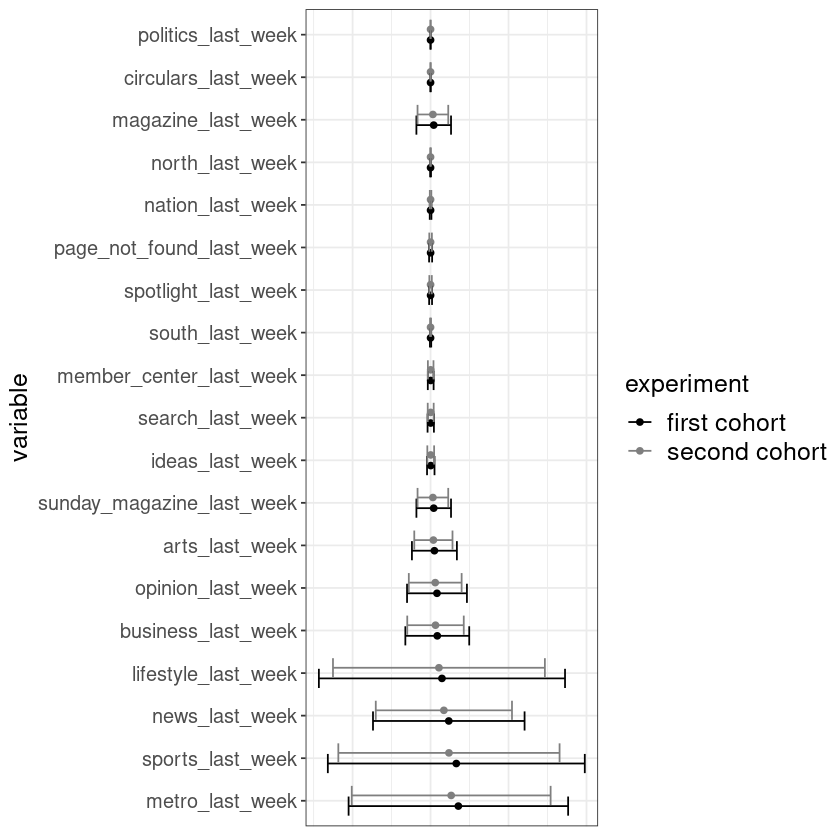}
    \caption{Content Consumption: Last Week}
  \end{subfigure}%
  \caption{Covariate shift: comparing the distribution (the two ends are 2.5 and 97.5 percentile) of continuous covariates (account activity and content consumption)}
  \label{fig17.1}
\end{figure}

\begin{figure}[!htb]
\centering
  \begin{subfigure}[t]{.6\textwidth}
    \centering
    \includegraphics[width=\linewidth]{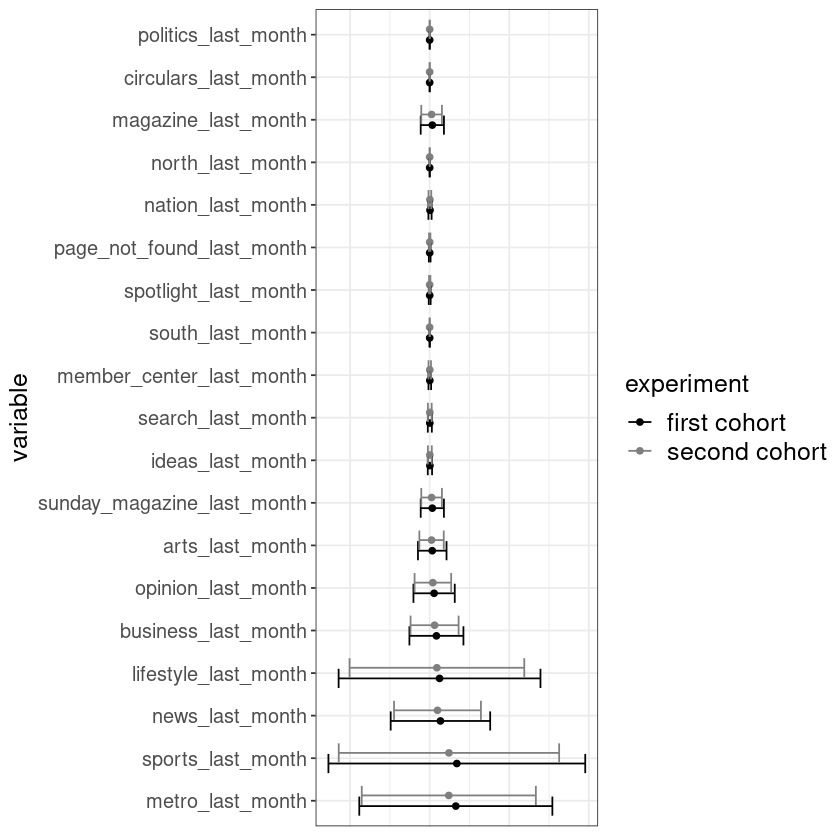}
    \caption{Content Consumption: Last Month}
  \end{subfigure}
  \hfill
  \begin{subfigure}[t]{.6\textwidth}
    \centering
    \includegraphics[width=\linewidth]{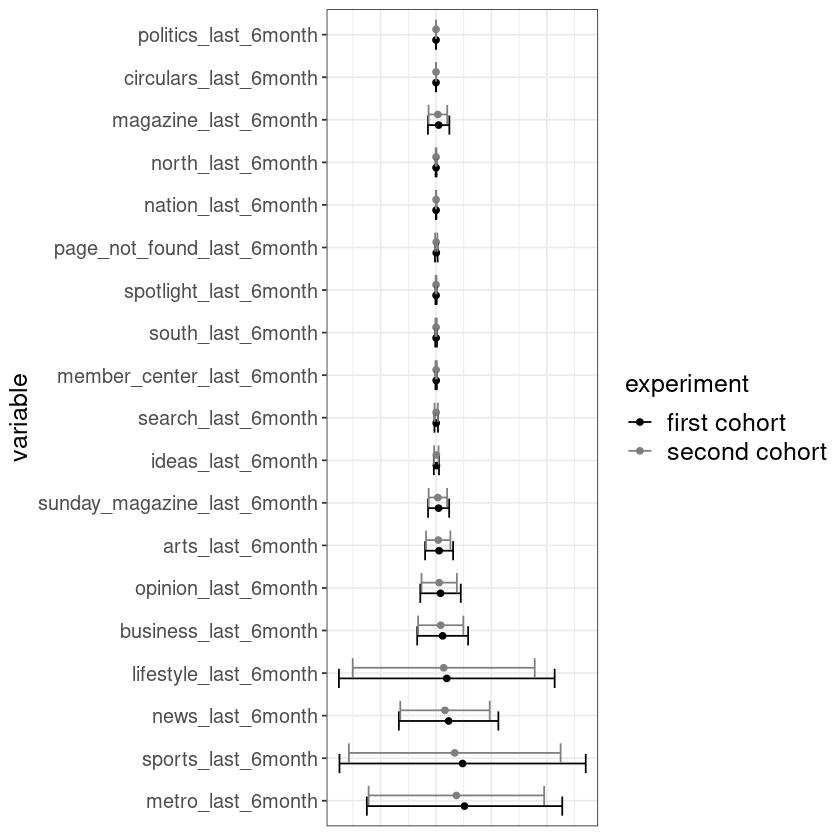}
    \caption{Content Consumption: Last 6 Month}
  \end{subfigure}
  %\bigskip
  \caption{Covariate shift: comparing the distribution (the two ends are 2.5 and 97.5 percentile) of continuous covariates (content consumption)}
  \label{fig17.2}
\end{figure}

Covariate shift means the distribution of subscriber features are quite different between the two experiments. When this is the case, the policy learned on the first experiment might not perform well on the second experiment because we are likely facing a different population. However, this doesn't seem to be the case in our data. Figure \ref{fig17.1} and \ref{fig17.2} show the distribution of covariates in the two experiments and we can see that they are quite similar. 

\begin{figure}[!htb]
    \centering
    \includegraphics[scale=0.7]{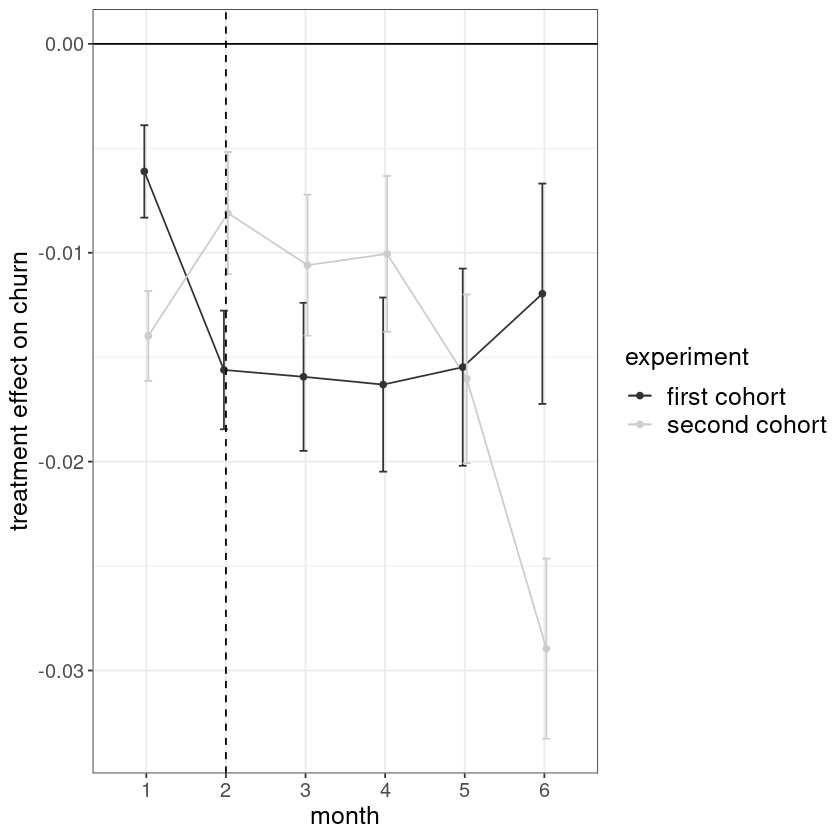}
    \caption{Concept shift: comparing the ATT overtime for two experiments. This is the treatment effect of the condition \$4.99/8 weeks relative to the control. We can only compare this condition because this is the only common treatment condition between the two experiments. The 95\% confidence intervals overlap for most of the time periods but month 1 and 6 are quite different.} 
    \label{fig18}
\end{figure}

Then we look at concept shift which is the change in relationship between outcome of interest, covariates and actions. We focus on the treatment effect here. Due to logistical constraints, we only have one common treatment between the two experiments, i.e., \$4.99/8 weeks. We plot the treatment effect over time from both experiments. Because we know the two populations are comparable in terms of observed covariates, the difference in treatment effect can be attributed to concept shift. The result is shown in Figure \ref{fig18}. We can see that the treatment effect over time looks somewhat different, so when learning the policy for future subscribers we only use data from the second experiment. Alternatively, we can pool data from both experiments but assign lower weights to observations in the first experiment to reflect the fact that this data is somewhat stale \citep{russac2019weighted}. 


\clearpage
\newpage
\subsection{Surrogate Choice}
\label{surrogate_choice}
\begin{figure}[h]
    \centering
  \includegraphics[scale=.45]{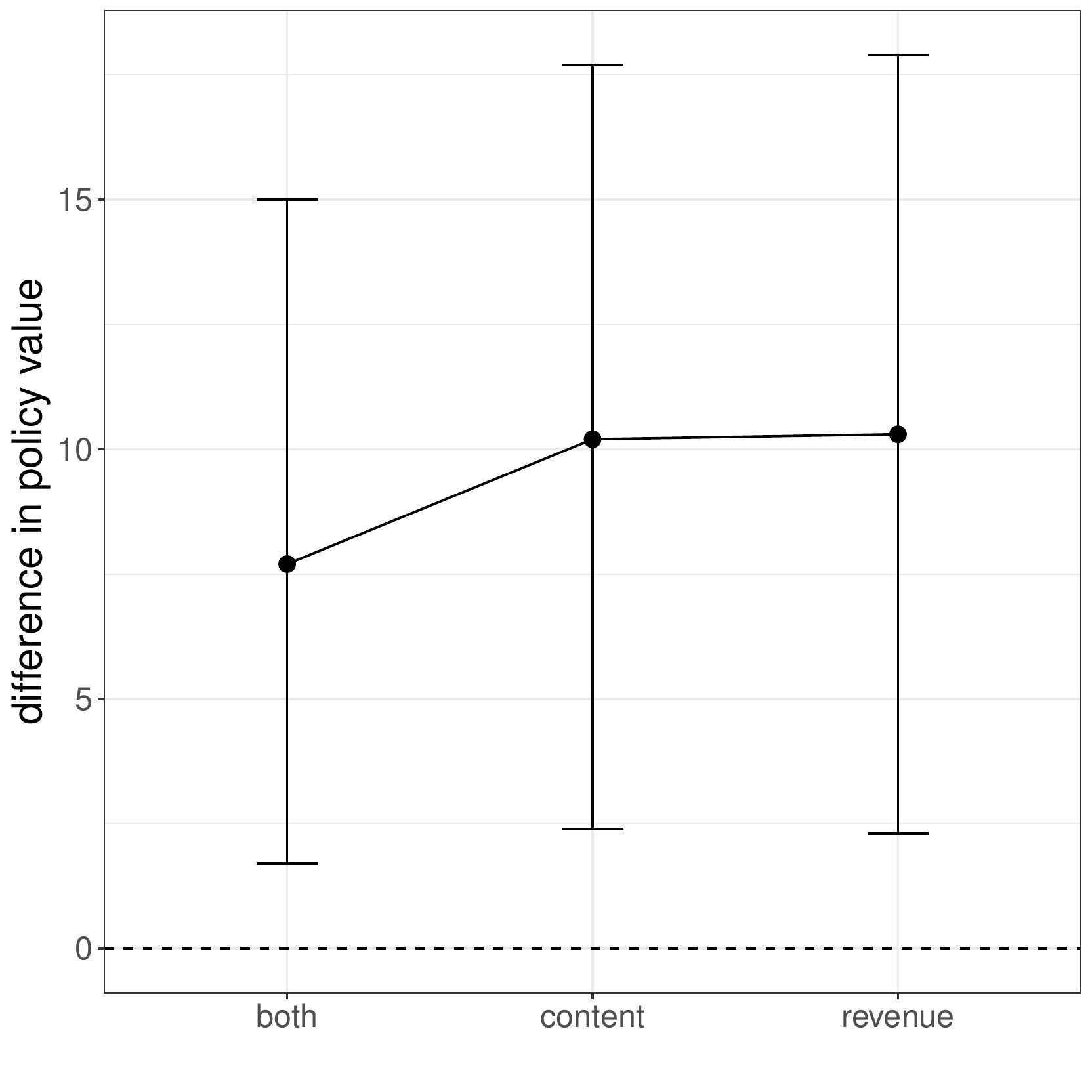}
    \caption{The value difference between policies learned with surrogate indices using content consumption variables, short-term revenue variables, or both, and the current policy. Each improves over the status quo.}
    \label{surrogate5}
\end{figure}

\clearpage
\newpage
\subsection{Power Simulation}

Before running the first experiment we conducted power simulation to see if we have enough power to detect any difference between alternative targeting policies. And we suspected that given the small number of treated subscribers our experiment might be under-powered. 

We vary two parameters: $q$, the percentage of subscribers targeted under the model and $\tau$, the effect size. For example, $q=0.01, \tau=0.1$ means that under the model we target the top 1\% of subscribers and the discount will lower the targeted subscribers' churn risk by 10\%. $Y(0)$, the outcomes without treatment are observed in the data, which is whether a given subscriber churned (churn = 1, not churn = 0). We simulate $Y(1)$ in the following way: for any subscriber whose $Y(0)$ is 0, we assume that the treatment won't \textit{increase} the churn risk so her $Y(1)$ is also 0. For any subscriber whose $Y(0)$ is 1, we flip a coin, with probability $1-\tau$ it stays 1 and with probability $\tau$ it becomes 0, $\tau$ is the effect size. After simulating the full schedule of potential outcomes we use the design policy discussed in Section \ref{sec:second_cohort} to simulate treatment assignment. The treatment assignment determines, for each individual, which potential outcome is revealed to us. This is considered one simulated experiment. Then for a fixed value of $q$ and $\tau$ and a full schedule of potential outcomes, we repeat the simulated experiment 100 times and calculate the power (percentage of simulated experiments that have a significant result) of different estimators. We look at both churn rate and implied revenue as our outcome measure. 

%and the results are summarized in Tables \ref{tab18} - \ref{tab21}

We find that for ATT using both churn rate and revenue as outcome, we have over 80\% of power only when the effect size is bigger than 20\%. And for ATT under model based targeting, we also need the effect size to be bigger than 20\% for 80\% power. For ATT under random targeting we will need even a bigger effect size at 30\%. We also calculated the total gain and loss for the campaign under the design policy and what it would be if we were to target using model based policy. We'd expect gains by using model based policy when effect size is moderately big (over 25\%) and we don't target too many subscribers (1 or 2\%). It turns out that our ATT is -28\%, it's within the range of $\tau$ that we covered in the simulation and bigger than we'd expected.

\end{appendices}
\end{document}